\pdfoutput=1
\documentclass{article}

\usepackage[english]{babel}

\usepackage[letterpaper,top=2cm,bottom=2cm,left=3cm,right=3cm,marginparwidth=1.75cm]{geometry}

\usepackage[parfill]{parskip}
\newlength{\defbaselineskip}
\usepackage{svg}
\usepackage{amsmath}
\usepackage{graphicx}
\graphicspath{ {figures/} }
\usepackage[colorlinks=true, allcolors=blue]{hyperref}
\usepackage{authblk}

\usepackage[utf8]{inputenc} 
\usepackage[T1]{fontenc}    
\usepackage{url}            
\usepackage{booktabs}       
\usepackage{amsfonts}       
\usepackage{nicefrac}       
\usepackage{microtype}      
\usepackage{xcolor}         
\usepackage{amssymb}
\usepackage{enumitem}
\usepackage{amsthm}     
\usepackage{cleveref}   
\usepackage{wrapfig}
\usepackage{subcaption}
\usepackage{multirow}
\usepackage{algorithm}
\usepackage[noend]{algpseudocode}
\usepackage{tabulary}
\usepackage{csquotes}
\usepackage{bbm}
\usepackage{tabularx}

\usepackage[hyphenbreaks]{breakurl}

\usepackage[citestyle=authoryear-comp,sorting=nyt,maxbibnames=99,backend=biber]{biblatex}
\allowdisplaybreaks
\usepackage[symbol]{footmisc}
\renewcommand{\thefootnote}{\fnsymbol{footnote}}

\title{An Empirical Study of Mamba-based Language Models}

\author{Roger Waleffe$^{1,2}$\thanks{Work done as an intern at NVIDIA}$\;$ Wonmin Byeon$^{1}$ Duncan Riach$^{1}$  Brandon Norick$^{1}$\thanks{Correspondence to \texttt{bnorick@nvidia.com}}$\;$ \\ Vijay Korthikanti$^{1}$
Tri Dao$^{3,4}$ Albert Gu$^{5,6}$
Ali Hatamizadeh$^{1}$ Sudhakar Singh$^{1}$\\
Deepak Narayanan$^{1}$ Garvit Kulshreshtha$^{1}$
Vartika Singh$^{1}$ Jared Casper$^{1}$ \\ Jan Kautz$^{1}$ Mohammad Shoeybi$^{1}$ Bryan Catanzaro$^{1}$}
\affil{$^{1}$NVIDIA \hspace{0.25cm} $^{2}$University of Wisconsin-Madison \hspace{0.25cm} $^{3}$Princeton University \\ $^{4}$Together AI \hspace{0.25cm} $^{5}$Carnegie Mellon University \hspace{0.25cm} $^{6}$Cartesia AI}

\date{}

\newcommand{\citep}{\parencite}
\newcommand{\citet}{\textcite}
\newcommand{\newparagraph}[1]{\vspace{1.5ex}\noindent\textbf{#1\hspace{1em}}}

\begin{document}
\maketitle

\begin{abstract}
\renewcommand{\thefootnote}{\arabic{footnote}}
\makeatletter
\renewcommand{\@makefntext}[1]{%
    \parindent 1em%
    \noindent
    \hb@xt@1.8em{\hss\@makefnmark}#1}
\makeatother
Selective state-space models (SSMs) like Mamba~\citep{gu2023mamba} overcome some of the shortcomings of Transformers, such as quadratic computational complexity with sequence length and large inference-time memory requirements from the key-value cache.
Moreover, recent studies have shown that SSMs can match or exceed the language modeling capabilities of Transformers, making them an attractive alternative.
In a controlled setting (e.g., same training data), however, studies so far have only presented small scale experiments (training with $<$3B parameters and $<$1T tokens) comparing SSMs to equivalent Transformers.
To understand the strengths and weaknesses of these architectures at larger scales, we present a direct comparison between 8B-parameter Mamba, Mamba-2, and Transformer models trained on the same datasets of up to 3.5T tokens. 
We also compare these models to an 8B-parameter hybrid architecture consisting of 43\% Mamba-2, 7\% self-attention, and 50\% MLP layers (Mamba-2-Hybrid).
Using a diverse set of natural language tasks, we answer the important question of whether Mamba models can match their Transformer counterparts at larger training budgets.
Our results show that while pure SSM-based models match or exceed Transformers on many tasks, both Mamba and Mamba-2 models lag behind Transformer models on tasks which require strong copying or in-context learning abilities (e.g., five-shot MMLU, Phonebook Lookup) or long-context reasoning.
In contrast, we find that the 8B-parameter Mamba-2-Hybrid exceeds the 8B-parameter Transformer on all 12 standard tasks we evaluated (+2.65 points on average) and is predicted to be up to 8$\times$ faster when generating tokens at inference time.
To validate long-context capabilities, we provide additional experiments evaluating variants of the Mamba-2-Hybrid and Transformer extended to support 16K, 32K, and 128K sequence lengths.
On an additional 23 long-context tasks, the hybrid model continues to closely match or exceed the Transformer on average.
To enable further study, we release the checkpoints as well as the code used to train our SSM-based models as part of NVIDIA's Megatron-LM project (\href{Megatron}{https://github.com/NVIDIA/Megatron-LM})\footnote[1]{A fixed snapshot of the code used in this technical report is available at\newline \hspace*{1.8em}\href{ssm}{https://github.com/NVIDIA/Megatron-LM/tree/ssm/examples/mamba}.}.

\end{abstract}

\section{Introduction}

Transformer-based large language models (LLMs)~\citep{vaswani2017attention} have become the dominant neural network architecture for natural language processing and have achieved impressive results across a wide array of tasks~\citep{achiam2023gpt, touvron2023llama}.
Much of the success of these models can be attributed to their self-attention layers~\citep{bahdanau2014neural}, which enable all-to-all information routing between tokens in a sequence, and their ability to improve with scaling model and dataset sizes.
However, self-attention layers suffer from some drawbacks that make training and deploying these models on long sequences challenging.
At training time, the computation required for self-attention layers scales quadratically with the sequence length.
At inference time, generating one token requires a memory capacity that scales linearly with the number of preceding tokens, necessitating a large key-value cache to store the required state.
Many recent works have attempted to address the efficiency issues with self-attention layers~\citep{tay2022efficient}; these works however have yet to match self-attention's language modeling capabilities.

Structured state space models~\citep{gu2021efficiently}, in particular Mamba~\citep{gu2023mamba} and more recently Mamba-2~\citep{dao2024transformers}, have been proposed as a promising alternative to self-attention layers and Transformers.
These models use constant computation and memory to generate a single token at inference time (after initializing the SSM states based on the context) and can be computed efficiently using hardware-aware algorithms during training.
They have been shown to match or exceed the downstream accuracy of Transformers on standard language modeling tasks for models up to 2.8B parameters~\citep{gu2023mamba, dao2024transformers}.
Follow up work has sought to further probe the in-context learning abilities of these models at small scale~\citep{park2024can}, and some recent work has investigated combining Mamba layers with attention layers to form \textit{hybrid} models~\citep{glorioso2024zamba, lieber2024jamba}. These works scale Mamba-based hybrid models beyond 7B parameters and show that doing so can result in high quality models.
However, in these studies the larger models were not compared with equivalent Transformers in a controlled setting (i.e., same training data, parameter count).
Such controlled comparisons have been limited to small-scale experiments and larger-scale studies of Mamba-2 models are still lacking.

In this technical report, we present a direct comparison between Mamba-based and Transformer-based LLMs trained on large datasets.
In particular, our primary goal is to provide a rigorous apples-to-apples comparison between Mamba, Mamba-2, Mamba-2-Hybrid (containing Mamba-2, attention, and MLP layers), and Transformers for 8B-parameter models trained on up to 3.5T tokens, with the same hyperparameters.
Using a diverse set of natural language tasks, we answer the important question of whether Mamba models can match their Transformer counterparts at larger training budgets.
We evaluate these models on 35 popular downstream language modeling tasks and use the exact same evaluation setup for Mamba-based and Transformer models. To ensure our evaluations are standard and reproducible, we provide details about the specific open-source benchmark suites and versions used in our experiments in Section~\ref{sec:preliminaries}.
Overall, our experiments eliminate the common difficulty of comparing LLMs, where it is often the case that both the model architecture but also the training data, tokenizer, and evaluation pipeline have changed.

Our experiments show that while Mamba and Mamba-2 models are good at modeling language (e.g., they match or exceed Transformers on many downstream tasks), they lag behind Transformer models when it comes to in-context learning and recalling information from the context. This confirms recent findings at smaller scales~\citep{park2024can}. In particular, we highlight the difficulty pure SSM models face with the standard five-shot MMLU~\citep{hendrycks2020measuring} and two-shot Phonebook tasks. For the former, after training for 1.1T tokens, both Mamba and Mamba-2 models produce nearly 15 points lower accuracy when compared to a Transformer model on this task. While the MMLU accuracy gap is partially addressed by training with more tokens (e.g., 3.5T tokens), SSM models still lag behind Transformer models for this common benchmark. We find that Phonebook and standard long-context benchmark tasks remain challenging for SSM models regardless of the number of training tokens.

Based on the above findings, we study in detail the potential for hybrid SSM-Transformer models to overcome the challenges faced by pure SSM architectures while retaining (some of) their inference-time benefits. Similar to \cite{lieber2024jamba}, we focus on LLMs consisting of a mixture of Mamba-2, self-attention, and MLP layers. Our ablation experiments aiming to identify the best hybrid model architecture lead us to design an 8B-parameter Mamba-2-Hybrid with 24 Mamba-2 layers, 4 self-attention layers, and 28 MLP layers. The self-attention and MLP layers are evenly distributed throughout the model. Extensive evaluations of this architecture show that it matches or exceeds Transformers on common natural language evaluations. When training for 3.5T tokens, a Mamba-2-Hybrid model exceeds a corresponding Transformer on all 12 short-context benchmarks we evaluated. On MMLU, the hybrid model reaches a five-shot accuracy 3.5 points higher than the Transformer. 

We also study long-context extensions of Mamba-2-Hybrid and the corresponding Transformer to support 16K and 32K context lengths. On 23 long-context evaluations, the 16K and 32K models closely match or exceed the Transformer baselines on average. Our results show that the hybrid models are particularly good at retrieving, tracking, and aggregating information over long contexts. We highlight three multi-document question answering tasks, however, which challenged the long-context hybrid models. We discuss potential reasons for these results and highlight areas of future work related to extending hybrid SSM-Transformer models to long sequence lengths.

Finally we highlight that, due to our use of global attention without any explicit position encoding in these models, long-context Mamba-2-Hybrid models can generalize beyond their trained sequence length. This is in contrast with recent hybrid models that use windowed attention and exhibit accuracy degradation on contexts larger than the window size but less than the pretraining sequence length~\citep{de2024griffin}. We find that a Mamba-2-Hybrid extended to support 128K contexts can perform the Phonebook lookup task perfectly even when the phone book contains more than 150K tokens.

We present our findings above to highlight the promise for larger-scale SSM-based models to provide faster, more efficient language model inference without compromising training efficiency or model accuracy compared to Transformers.
We hope that by releasing these results, the community is further excited by the potential of Mamba-based LLMs.
To help enable further adoption, we release the code used to train our Mamba, Mamba-2, and Mamba-2-Hybrid hybrid models as part of NVIDIA's Megatron-LM library (\href{Megatron}{https://github.com/NVIDIA/Megatron-LM}).
We also release the model weights for our Mamba-2 8B and Mamba-2-Hybrid 8B on Hugging Face.

\section{Preliminaries}
\label{sec:preliminaries}
In this section, we discuss briefly our implementation of SSM layers in Megatron-LM and discuss the training data and evaluations used throughout this report.

\begin{figure}[t]
  \centering
    \centering
    \includegraphics[width=0.95\textwidth]{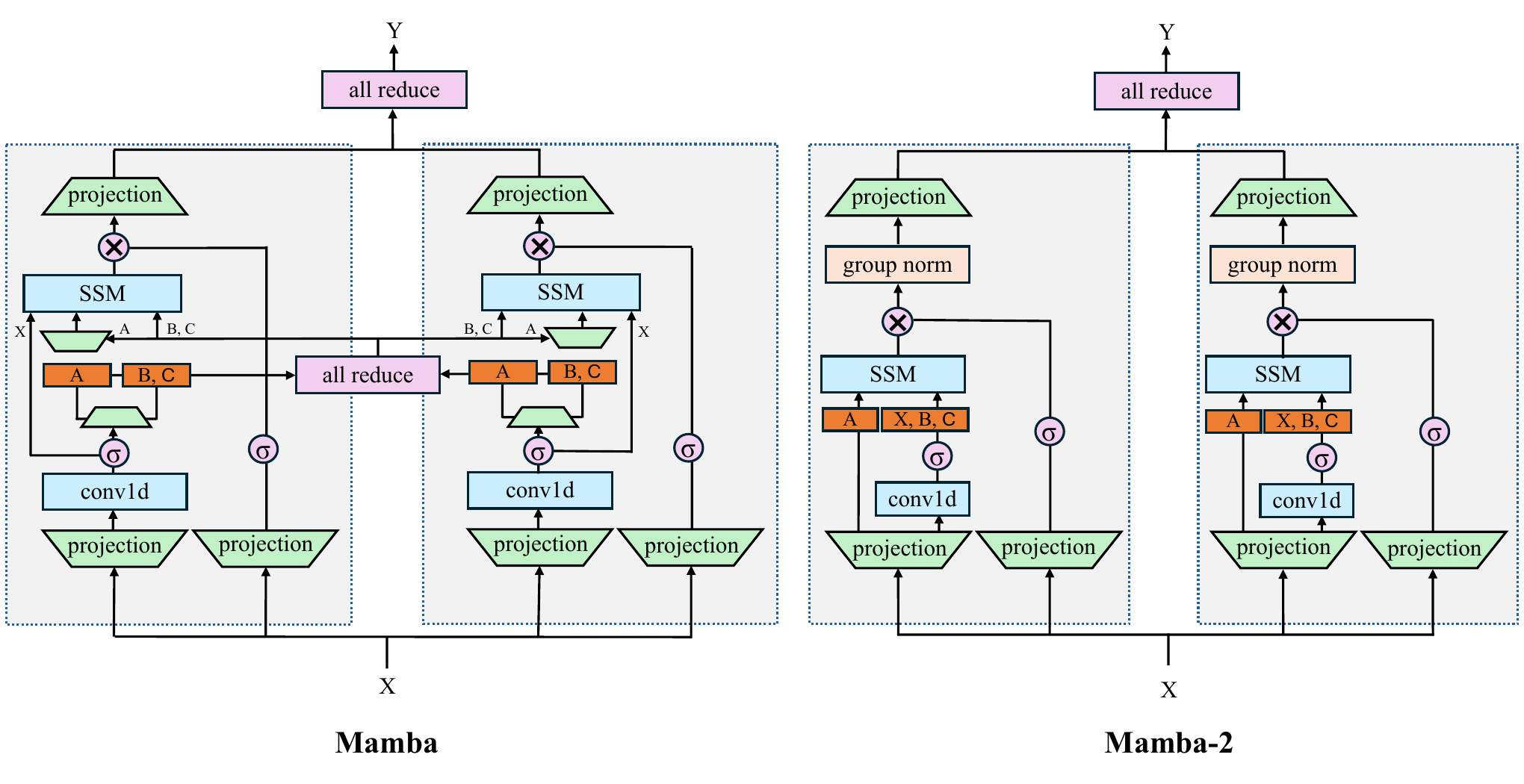}
  \caption{Mamba and Mamba-2 blocks with tensor model parallel size two. Mamba requires two all-reduces per layer while Mamba-2 requires only one. More details can be found in~\cite{dao2024transformers}.}
  \label{fig:model}
\end{figure}
\subsection{Model Implementation}
To support efficient large-scale training, we implement Mamba and Mamba-2 layers with support for tensor~\citep{shoeybi2019megatron}, sequence~\citep{korthikanti2205reducing}, and pipeline parallelism~\citep{narayanan2021efficient} (only for Mamba-2). As described in~\cite{dao2024transformers}, tensor-parallel support for Mamba layers requires two all-reduces per block compared to just one all-reduce for Transformer layers (Figure~\ref{fig:model}), leading to increased communication overheads for training larger-scale Mamba models. Mamba-2 tensor parallel support, on the other hand, requires only one all-reduce per layer, but requires the use of GroupNorm rather than LayerNorm for the internal block normalization (see Figure~\ref{fig:model}).

We found that using GroupNorm lead to no difference in validation loss when compared to using full LayerNorm as long as the group size (the model hidden dimension divided by the number of groups) is sufficiently large to allow for accurate calculations of the per-group normalization statistics (in our experience this meant a group size greater than 256). To implement SSM-Transformer hybrid models, we combine our Mamba or Mamba-2 layers with the existing self-attention and MLP layers supported in Megatron-LM. These layers support all the previously mentioned parallelization strategies enabling us to immediately train hybrid models with tensor, sequence, and pipeline parallelism.

\subsection{Training Data}
We train the models discussed in this report on 1.1T and 3.5T token datasets. Both datasets are predecessors of the dataset used to train Nemotron-4 and are comprised of 70\% English, 15\% non-English, and 15\% code. For additional details, refer to the discussion included in the Nemotron-4 technical report~\citep{parmar2024nemotron}. We use a vocabulary of 256K tokens trained with SentencePiece~\citep{kudo2018sentencepiece}.

\subsection{Evaluation Tasks and Setup}
We now discuss the evaluations used throughout the paper. Wherever possible, we use open-source LLM benchmark suites to ensure our evaluations are standard and reproducible. We report results using a large number of common tasks:

\begin{itemize}
    \item \textbf{Standard Short-Context Tasks:} We use the open-source LM Evaluation Harness library (commit \texttt{94cc1850})~\citep{eval-harness} to evaluate the following 12 tasks (metric used for evaluation reported in parentheses): WinoGrande (accuracy)~\citep{sakaguchi2021winogrande}, PIQA (accuracy)~\citep{bisk2020piqa}, HellaSwag (normalized accuracy)~\citep{zellers2019hellaswag}, ARC-Easy and ARC-Challenge (accuracy and normalized accuracy)~\citep{clark2018think}, MMLU (accuracy)~\citep{hendrycks2020measuring}, OpenBookQA (normalized accuracy)~\citep{mihaylov2018can}, TruthFulQA (accuracy)~\citep{lin2021truthfulqa}, PubMedQA (accuracy)~\citep{jin2019pubmedqa}, and RACE (accuracy)~\citep{lai2017race}. Each of the proceeding tasks are evaluated by measuring the probability returned by the model for each possible answer choice. We also use the generation-based tasks Natural Questions (NQ) (exact match)~\citep{lee2019latent} and SquadV2 (F1)~\citep{rajpurkar2018know}.
    
    \item \textbf{Natural Long-Context Tasks:} To evaluate long-context models, as above, we use three tasks from LM Evaluation Harness: NarrativeQA (F1)~\citep{kovcisky2018narrativeqa}, Qasper (F1)~\citep{dasigi2021dataset}, and QuALITY (normalized accuracy)~\citep{shaham2022scrolls}. The first two tasks are generation-based, while the latter uses continuation probabilities returned by the model for each answer. Each of these three tasks requires the model to answer a given question based on a long input document. 
    
    We also use six tasks from the LongBench~\citep{bai2023longbench} long-context evaluation benchmark (commit \texttt{48798083}): MultiFieldQA-English (F1), HotpotQA (F1)~\citep{yang2018hotpotqa}, 2WikiMQA (F1)~\citep{ho2020constructing}, Musique (F1)~\citep{trivedi2022musique}, TREC (accuracy)~\citep{li2002learning}, and TriviaQA (F1)~\citep{joshi2017triviaqa}. Each of these six tasks requires the model to generate the answer. MultiFieldQA tests a model's ability to perform single-document question answering while HotpotQA, 2WikiMQA, and Musique measure multi-document question answer capabilities. TREC and TriviaQA are used to measure a model's ability to perform in-context learning over long inputs.
    
    \item \textbf{Synthetic Long-Context Tasks:} Finally, we also evaluate our models using synthetic tasks that aim to measure a model's ability to retrieve, track, and aggregate information across long input texts. For these evaluations, we use the Phonebook task introduced in \cite{jelassi2024repeatafterme} (illustrated in Figure~\ref{fig:phonebook_task}) and 13 open-source, generation-based tasks in the RULER benchmark, described explicitly in Appendix B of~\cite{hsieh2024ruler}. The RULER tasks consist of eight Needle In A Haystack (NIAH) variants, one multi-hop tracing task called Variable Tracking (VT), two long-context aggregation tasks (Common Words Extraction (CWE) and Keywords Extraction (KWE)), one single-document question answer task (SquadQA), and one multi-document question answer task (HotpotQA). For all tasks, we report the accuracy on 400 synthetic samples generated by RULER.
\end{itemize}
\section{Mamba and Mamba-2 Compared to Transformers}
In this section we discuss our observations and experimental results training 8 billion (8B) parameter Mamba and Mamba-2 models and compare with 8B-parameter Transformer models. We find that Mamba and Mamba-2 can match or exceed Transformers on standard zero-shot tasks (Section \ref{subsec:downstream_eval}) but lag behind on MMLU and copying tasks, which we discuss in details in Sections~\ref{subsubsec:mmlu_in_detail} and~\ref{subsec:pure_model_copying}.

\begin{table}[t]
\caption{8B-parameter Mamba, Mamba-2, and Transformer architectures used in the experiments.}
\label{tab:pure_model_architectures}
\centering
\resizebox{\textwidth}{!}{
\begin{tabular}{cccccccccc}
\toprule
Model & Params (B) & \# Layers & Model Dim & Attn. Heads & State Dim. & \# Groups & Pos. Emb. & Seq. Len\\
\midrule
Transformer & 8.53 & 32 & 4096 & 32 & - & - & Rope & 4096\\
Mamba & 8.15 & 56 & 4096 & - & 128 & - & None & 4096\\
Mamba-2 & 8.24 & 56 & 4096 & - & 128 & 8 & None & 4096\\
\bottomrule
\end{tabular}
}
\end{table}

\subsection{Model Architectures}
\label{subsec:pure_model_arch}
We train Mamba, Mamba-2, and Transformer models with the architectures summarized in Table~\ref{tab:pure_model_architectures}. We discuss the architectures in more detail next. Additional details can be found in the released model checkpoints and open-sourced code in Megatron-LM.

\newparagraph{Transformer.}
Our 8B Transformer model follows the style of GPT3~\citep{brown2020language} and consists of 32 layers (each Multi-Head Attention + MLP) with a hidden dimension of 4096. We use 32 attention heads, 128 KV-channels, a $4\times$ expansion for the MLPs, SwiGLU activation~\citep{shazeer2020glu}, LayerNorm~\citep{ba2016layer}, and RoPE~\citep{su2024roformer} for position embeddings. We do not use bias weights for linear layers or Dropout. Additionally, we use seperate parameters for model embeddings and output layer weights (which we refer to as \textit{untied embeddings}).

\newparagraph{Mamba.} We train an 8B-parameter Mamba model with hidden dimension 4096 and 56 layers (typically 2 Mamba layers have around the same parameters as one block of attention + MLP). The state dimension for each Mamba layer is set to 128 and we use GELU~\citep{hendrycks2016gaussian} activation. Following~\citep{gu2023mamba}, we do not use any explicit position encoding and for normalization we use RMSNorm~\citep{zhang2019root}. As for the Transformer, we do not use bias weights for linear layers or Dropout and we use untied embeddings. 

\newparagraph{Mamba-2.} For Mamba-2 models, we use the same architecture as above for Mamba except replace each layer with the updated Mamba-2 block~\citep{dao2024transformers}. We set the internal Mamba-2 state dimension to 128 and use eight groups. We retain the default values from~\cite{dao2024transformers} and use a head dimension of 64, expansion factor of two, and window size of four for convolution.

\subsection{Training Hyperparameters}
We train the above models on 1.1T and 3.5T token datasets (see details in Section~\ref{sec:preliminaries}) using the following hyperparameters: On the smaller dataset, we use a batch size of 256, peak learning rate of 1e-4 and minimum learning rate of 1e-5. On the larger dataset we increase the batch size to 1024 and use higher learning rates: a peak of 3e-4 and minimum of 3e-5. On both datasets we use learning rate warm up over 122K samples, a cosine learning rate schedule, weight decay of 0.1, 0.9 and 0.95 for Adam $\beta_1$ and $\beta_2$ parameters respectively, and train using BF16. We performed some studies at smaller scale and found that Mamba network hyperparameters are similar to that of Transformers and as a result we use the same hyperparameters across models to make a rigorous direct comparison.

\begin{table}[t]
\caption{Evaluation results for 8B-parameter models trained on 1.1T tokens. Pure SSM models (Mamba and Mamba-2) match or exceed Transformers on many natural language tasks, but fall short on others (e.g., MMLU) (see Section~\ref{subsec:pure_model_eval}).} 
\label{tab:one_t_pure_models}
\centering
\resizebox{\textwidth}{!}{
\begin{tabular}{cccccccccc}
\toprule
\multirow{2}{*}{Model} & \multirow{2}{*}{WinoGrande} & \multirow{2}{*}{PIQA} & \multirow{2}{*}{HellaSwag} & \multirow{2}{*}{ARC-E} & \multirow{2}{*}{ARC-C} & \multicolumn{2}{c}{MMLU} & \multirow{2}{*}{Avg. w/o MMLU} & \multirow{2}{*}{Avg}\\
& & & & & & 0-Shot & 5-Shot & & \\
\midrule
Transformer & 69.22 & 78.29 & 75.6 & 73.15 & \textbf{43.09} & \textbf{38.32} & \textbf{46.28} & 67.87 & \textbf{60.56} \\
\midrule
Mamba & 68.27 & \textbf{78.89} & \textbf{75.63} & \textbf{75.42} & 42.15 & 28.63 & 28.00 & 68.07 & 56.71 \\
Mamba-2 & \textbf{70.8} & 78.35 & 75.54 & 75.13 & 43.00 & 28.94 & 29.19 & \textbf{68.56} & 57.28 \\
\bottomrule
\end{tabular}
}
\end{table}

\subsection{Empirical Evaluation of Mamba and Mamba-2}
\label{subsec:pure_model_eval}

\subsubsection{Downstream Language Modeling Tasks}
\label{subsec:downstream_eval}
In Table~\ref{tab:one_t_pure_models} and~\ref{tab:three_t_pure_models} we report the results of training our 8B-parameter Mamba, Mamba-2, and Transformer models on 1.1T and 3.5T tokens respectively, using six standard tasks for measuring natural language understanding. On the 3.5T dataset, we train only a pure Mamba-2 model (and not a Mamba model) for efficiency reasons---the pure Mamba model on the 1.1T dataset was almost 3$\times$ slower than the pure Mamba-2 model due to the large state dimension. 

Our results confirm those of prior works~\citep{dao2024transformers}; both Mamba and Mamba-2 models can match or exceed Transformer models on common tasks. On both datasets, pure SSM models achieve higher accuracy than Transformers when averaged over the WinoGrande, PIQA, HellaSwag, ARC-Easy, and ARC-Challenge evaluation tasks. The results on the 1.1T dataset also highlight that pure Mamba-2 models are equal or better than Mamba models on average. The most interesting observation from these results, however, is that the accuracy on MMLU is significantly worse for pure SSM models compared to the Transformer when training for the shorter token horizon (1.1T tokens). For example Mamba-2 five-shot accuracy is 17 points lower than that of the Transformer in this setting. Table~\ref{tab:three_t_pure_models} shows that training for more tokens helps the Mamba-2 model improve on MMLU (visualized in Figure~\ref{fig:mmlu_vs_training}), closing the gap to the Transformer to just 1.37 points. We discuss the MMLU result in more detail in the following section.

\begin{table}[h]
\caption{Evaluation results for 8B-parameter models trained on 3.5T tokens. We train only pure Mamba-2 and Transformer models in this setting due to efficiency issues with training larger-scale Mamba models. Training for 3.5T tokens instead of 1.1T (Table~\ref{tab:one_t_pure_models}) allows Mamba-2 to approach Transformer-level accuracy on MMLU and produces a pure SSM model that exceeds the Transfomer in terms of average task accuracy.} 
\label{tab:three_t_pure_models}
\centering
\resizebox{\textwidth}{!}{
\begin{tabular}{cccccccccc}
\toprule
\multirow{2}{*}{Model} & \multirow{2}{*}{WinoGrande} & \multirow{2}{*}{PIQA} & \multirow{2}{*}{HellaSwag} & \multirow{2}{*}{ARC-E} & \multirow{2}{*}{ARC-C} & \multicolumn{2}{c}{MMLU} & \multirow{2}{*}{Avg. w/o MMLU} & \multirow{2}{*}{Avg}\\
& & & & & & 0-Shot & 5-Shot & & \\
\midrule
Transformer & 69.14 & 78.62 & 75.89 &73.27 & 43.77 & 45.69 & \textbf{50.07} & 68.14 & 62.35 \\
\midrule
Mamba-2 & \textbf{71.59} & \textbf{79.82} & \textbf{77.69} & \textbf{75.93} & \textbf{48.12} & \textbf{47.25} & 48.7 & \textbf{70.63} & \textbf{64.16} \\
\bottomrule
\end{tabular}
}
\end{table}

\subsubsection{A Closer Look at MMLU}
\label{subsubsec:mmlu_in_detail}
\begin{figure}[t]
  \centering
  \begin{subfigure}[t]{0.49\textwidth}
    \centering
    \includegraphics[width=\textwidth]{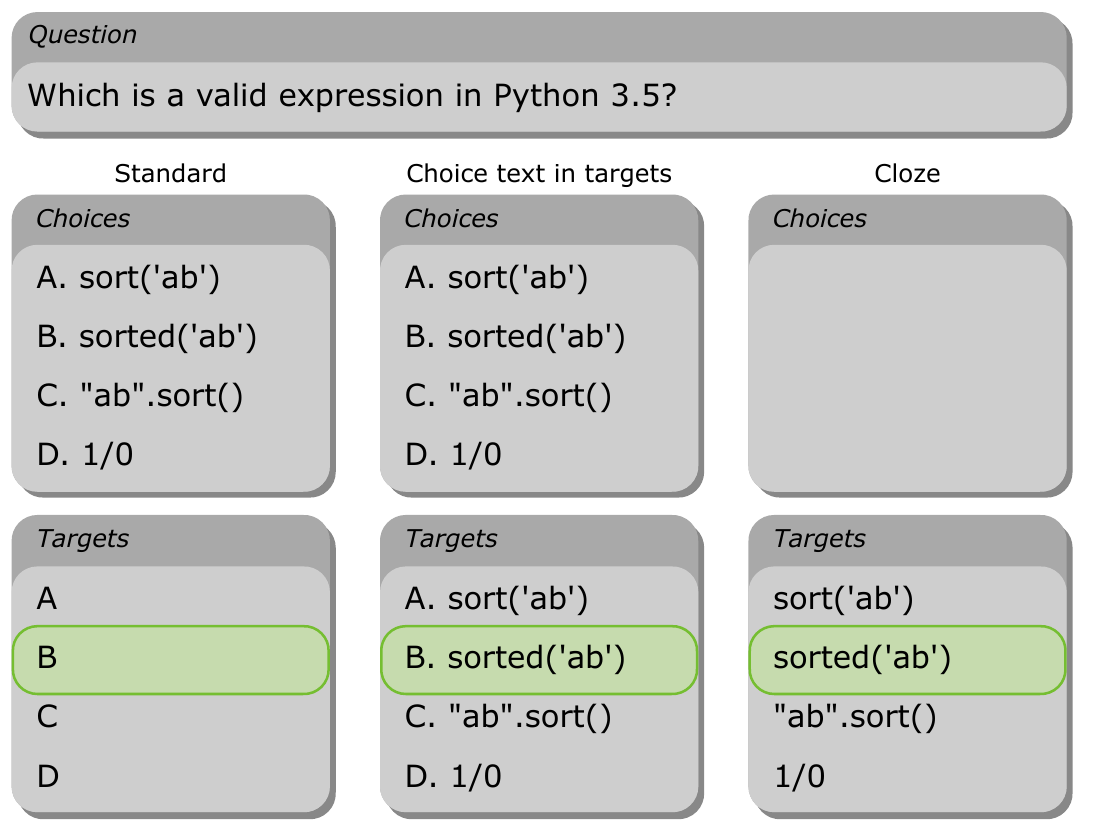}
    \caption{Three different formats for MMLU multiple choice questions. In all cases the model is first prompted with a question. In the standard and `choice text in targets' variants (but not `cloze' variant), the prompt includes four multiple choice answers following the question. The correct answer is then calculated by measuring which of the four answers, represented in the target format, is predicted to have the highest probability of following the given prompt.}
  \label{fig:mmlu_formats}
  \end{subfigure}
  \hfill
  \begin{subfigure}[t]{0.49\textwidth}
    \centering
    \includegraphics[width=\textwidth]{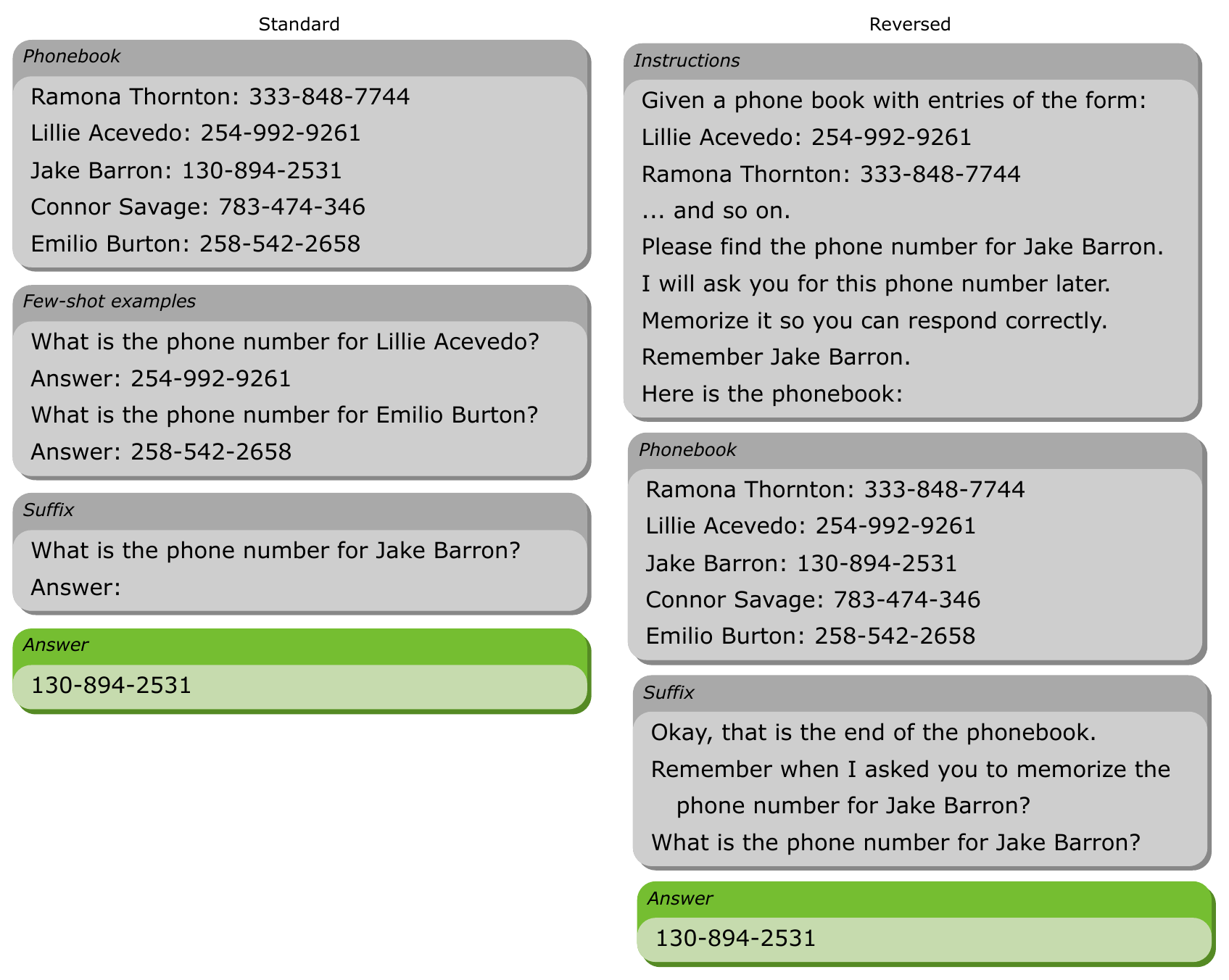}
    \caption{Two different versions of the synthetic Phonebook task. In the standard configuration (left) the model is first prompted with a phone book. It is then shown two in-context example questions before the final question which asks it to recall the phone number for a specific person. In the `reversed' setting (right) the model is told which phone number to look for before being shown the phone book and then later asked to recall this number.}
    \label{fig:phonebook_task}
  \end{subfigure}
  \caption{Illustrations of the MMLU and Phonebook tasks used in the experiments.}
  \label{fig:prompt_illustrations}
\end{figure}

We investigate the gap in MMLU accuracy between pure SSM models and Transformers by evaluating our 1.1T Mamba, Mamba-2, and Transformer models (where the gap is largest) on different instances of this task. Generally, MMLU accuracy is calculated by prompting the model with a question that includes four answer choices labeled with the letters A, B, C, and D. The model is then shown each of the four letters A, B, C, and D and the letter most likely to follow the prompt (measured by probabilities output from the model) is taken as the model's answer (Figure~\ref{fig:mmlu_formats}). MMLU accuracy, however, can also be measured by calculating the probability of the full answer choice following the prompt (which we call a \textit{choice-text-in-targets} variant) or using a \textit{cloze} format. In the latter case, the model is prompted with only the question (no answer choices are provided) and the text of each answer is used to calculate probabilities. 

\begin{table}[t]
\small
\caption{MMLU accuracy for pure SSM and Transformer models according to the three task variants described in Figure~\ref{fig:mmlu_formats}. Results are for 8B-parameter models trained on 1.1T tokens. While pure SSM models struggle with the standard and choice-text-in-targets formulations on this token horizon, they match or exceed the Transformer model in the cloze setting. Together with the results for Mamba-2 trained on 3.5T tokens (Table~\ref{tab:three_t_pure_models}, Figure~\ref{fig:mmlu_vs_training}), this suggests that SSM models contain the same knowledge as Transformers, but require more training to understand the MMLU task formatting.}
\label{tab:mmlu_default_vs_cloze}
\centering
\begin{tabular}{ccccccc}
\toprule
\multirow{2}{*}{Model} & \multicolumn{2}{c}{MMLU-Standard} & \multicolumn{2}{c}{MMLU-W/Choice} & \multicolumn{2}{c}{MMLU-Cloze} \\
& 0-Shot & 5-Shot & 0-Shot & 5-Shot & 0-Shot & 5-Shot \\
\midrule
Transformer & \textbf{38.32} & \textbf{46.28} & \textbf{33.54} & \textbf{46.64} & 37.26 & 39.24\\
\midrule
Mamba & 28.63 & 28.00 & 27.42 & 29.17 & \textbf{38.26} & \textbf{39.28}\\
Mamba-2 & 28.94 & 29.19 & 28.54 & 30.68 & 37.68 & 38.17 \\
\bottomrule
\end{tabular}
\end{table}

We show the results of evaluating our 8B-parameter pure SSM and Transformer models trained on 1.1T tokens on the three formulations of MMLU described above in Table~\ref{tab:mmlu_default_vs_cloze}. While the pure SSM models struggle with the standard and choice-text-in-targets formulations, they actually exceed the accuracy of the Transformer in the cloze setting. This experiment, together with the MMLU results for Mamba-2 trained on 3.5T tokens (Table~\ref{tab:three_t_pure_models}, Figure~\ref{fig:mmlu_vs_training}), highlight that the \textbf{pure SSM models contain the same knowledge as the Transformer, but that they require substantially more training to understand the format of the multiple choice questions} in the first two settings. 
We hypothesize that the reason for this confusion, especially in the standard MMLU setting, is that pure SSM models are unable to directly \textit{route} the knowledge of each answer into a single answer token. In contrast, the self-attention layers in the Transformer are particularly good at that task, especially when they are shown several in-context examples that teach them to do such routing (e.g., 5-Shot MMLU in the standard formulation). Finally, we note that while the Mamba-2 hybrid model trained for 3.5T tokens closes the MMLU gap to the Transformer, it sees an accuracy improvement on standard MMLU of only 1.45 points when moving from 0- to 5- shot, compared with 4.38 for the Transformer, providing additional evidence that Transformers may have superior in-context learning capabilities.

\begin{figure}
  \centering
  \begin{subfigure}[t]{0.32\textwidth}
     \centering
     \includegraphics[width=\textwidth]{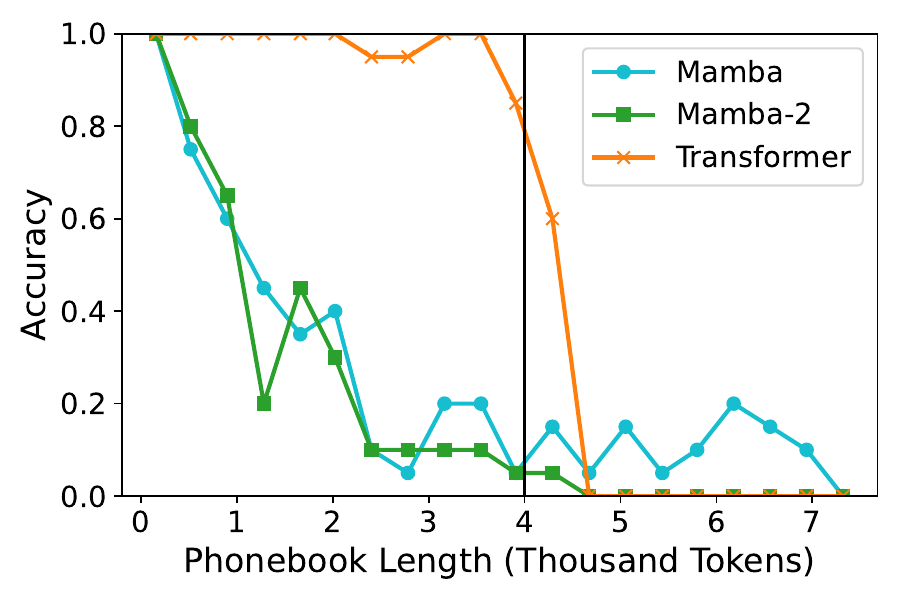}
     \caption{}
     \label{fig:phonebook_d_1t}
  \end{subfigure}
  \hfill
  \begin{subfigure}[t]{0.32\textwidth}
     \centering
     \includegraphics[width=\textwidth]{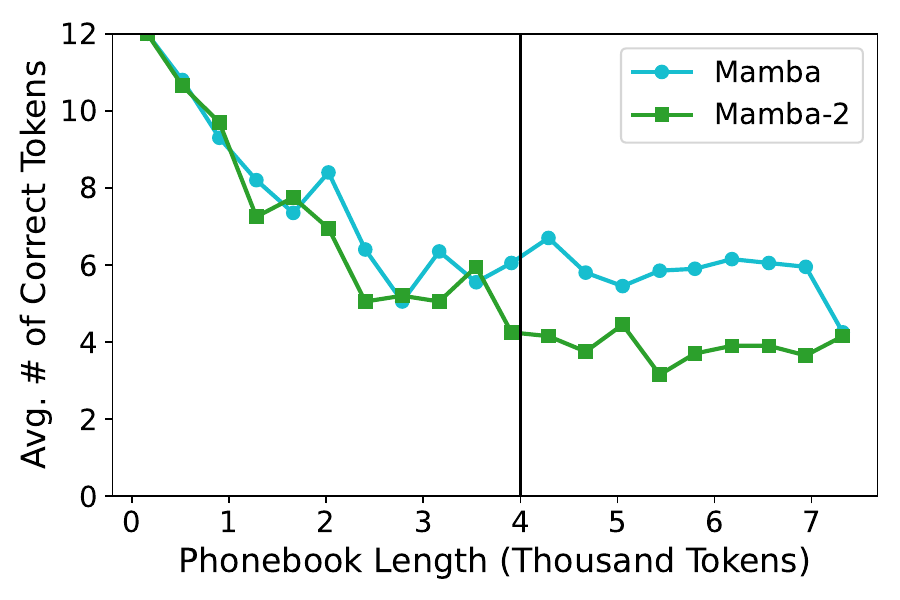}
     \caption{}
     \label{fig:phonebook_cd_1t}
  \end{subfigure}
  \hfill
  \begin{subfigure}[t]{0.32\textwidth}
     \centering
     \includegraphics[width=\textwidth]{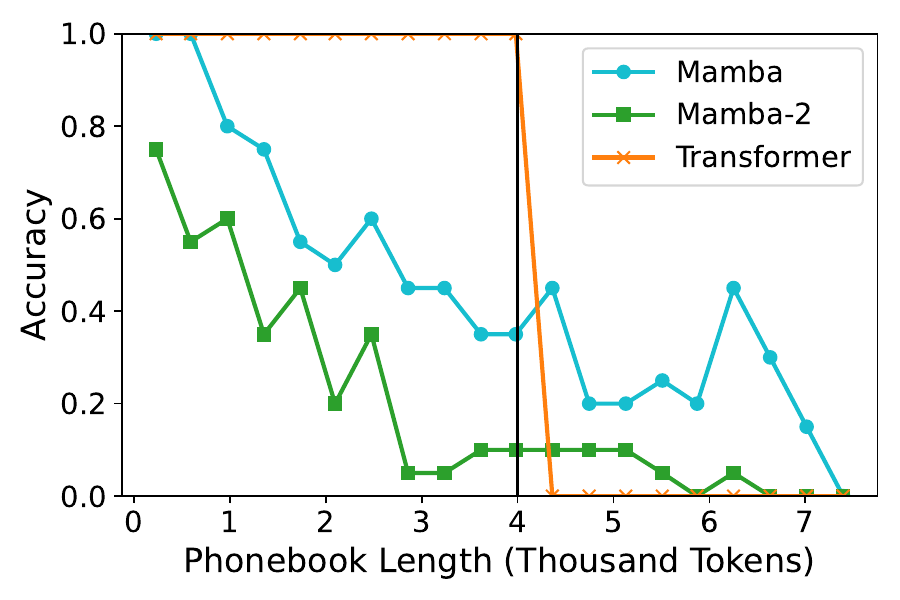}
     \caption{}
     \label{fig:phonebook_r_1t}
  \end{subfigure}
  \caption{Evaluation results for pure SSM and Transformer models (trained for 1.1T tokens) on the Phonebook task illustrated in Figure~\ref{fig:phonebook_task}. \textbf{(a)} On the standard Phonebook task, Transformers are capable of in-context learning and answering questions that require copying from the input, but SSM models struggle with this task. \textbf{(b)} In the standard Phonebook setting (i.e., (a)), SSM models exhibit \textit{fuzzy memory}---while they are unable to correctly predict the phone number, they predict phone numbers that share multiple digits (in the right locations) with the correct answer (see Section~\ref{subsec:pure_model_copying}). \textbf{(c)} On the Reversed Phonebook formulation, even when notified at the beginning of the context which phone number they will be asked to recall, SSM models still lag behind Transformer models.}
  \label{fig:phonebook_1t_plots}
\end{figure}

\subsubsection{Copying Tasks}
\label{subsec:pure_model_copying}

Beyond downstream language modeling tasks, we also evaluate pure SSM-based models and compare to Transformers on the synthetic Phonebook task \citep{jelassi2024repeatafterme} that aims to measure a model's ability to perform in-context learning (through few shot examples) and copying from earlier in the context. We illustrate an example Phonebook prompt in Figure~\ref{fig:phonebook_task}. The model is first prompted with a list of (name, phone number) pairs, and then asked `What is the phone number for \{name\}?' with two example question answer pairs before the actual question used for testing. For each trial, we randomly generate names and phone numbers to create the phone book and randomly select which names are used for the two examples and the final query. Accuracy on this task is then measured by whether the model generates the correct phone number or not.

We vary the length of the phone book (the number of (name, phone number) pairs) and plot the accuracy for each phone book length averaged over 20 different random initializations in Figure~\ref{fig:phonebook_d_1t}. The 8B Transformer model can respond correctly with near 100\% accuracy for phone book lengths up to its pretraining context length (4096). In contrast, both Mamba and Mamba-2 models begin to respond incorrectly for input sequence lengths beyond approximately 500 tokens. In contrast to MMLU, this behavior persists for Mamba-2 even when training for 3.5T tokens (Figure~\ref{subfig:phonebook_4k}).

A closer look at the SSM model predictions shows that while they cannot perfectly recall the correct phone number, these models \textit{have} compressed information about each phone book entry into their running states---we show in Figure~\ref{fig:phonebook_cd_1t} the average number of correct tokens predicted by Mamba and Mamba-2 on Phonebook by comparing the predicted answer to the true answer. 
Figure~\ref{fig:phonebook_cd_1t} shows that pure SSM-based models have \textit{fuzzy memory}. That is, while they cannot predict the phone number exactly, they do generally respond with phone numbers that are similar to the correct answer.

Finally, we evaluate whether changing the Phonebook prompt allows for SSM models to achieve better results. In particular, we prompt the model with the name of the person whose phone number it will be asked to recall \textit{before} showing it the phone book (the Reversed formulation in Figure~\ref{fig:phonebook_task}). 
Figure~\ref{fig:phonebook_r_1t} shows the results of the 8B Mamba, Mamab-2, and Transformer models in this modified Phonebook setting. Interestingly, while the SSM models achieve better accuracy as a function of phone book length using this prompt, the accuracy still degrades for phone books with lengths shorter than 4096 (the sequence length used for pretraining). Even with the modified Phonebook prompt, it remains challenging for the SSM to decide which information to store exactly and which information to forget on this task. We hypothesize that finetuning Mamba and Mamba-2 on the Phonebook task would lead to improved accuracy.

\subsubsection{Takeaway}
Our experiments training 8B-parameter Mamba and Mamba-2 models showed that while these models achieve comparable or better accuracy than Transformers on many standard natural language modeling tasks, they achieve lower accuracy on others. In particular, we identified MMLU (with smaller token horizons) and Phonebook as challenging tasks for pure SSM-based models and hypothesize that this is because these tasks require in-context learning, information routing between tokens, and copying from the context.

\section{Hybrid Mamba-Transformer Models}
Motivated by the difficulties pure SSM models face with retrieving information from the context and in-context learning, we now study the hypothesis that adding a few Transformer layers (made of self-attention and MLP layers) back into the architecture enables the model to overcome these issues. In this section we consider \textit{hybrid} models containing a combination of Mamba/Mamba-2, self-attention, and MLP layers.

\subsection{Designing a Hybrid SSM-Transformer Model}
\label{subsec:hybrid_design}
We begin by discussing the ablation studies that led us to design our final hybrid model architecture.  
For the experiments reported in this section, all model variants have the same number of parameters per layer. This ensures that model quality changes are not due to an increase or decrease in the overall number of parameters, and also that we can control the ratio of parameters by controlling the ratio of layers. To do so, we adjust both the number of attention heads (while keeping the head size constant) and the MLP expansion factor such that self-attention and MLP layers have approximately the same number of parameters as Mamba layers.

\begin{wrapfigure}{R}{0.4\textwidth}
\vspace{-0.2in}
\begin{minipage}{0.4\textwidth}
\centering
    \includegraphics[width=1.0\linewidth]{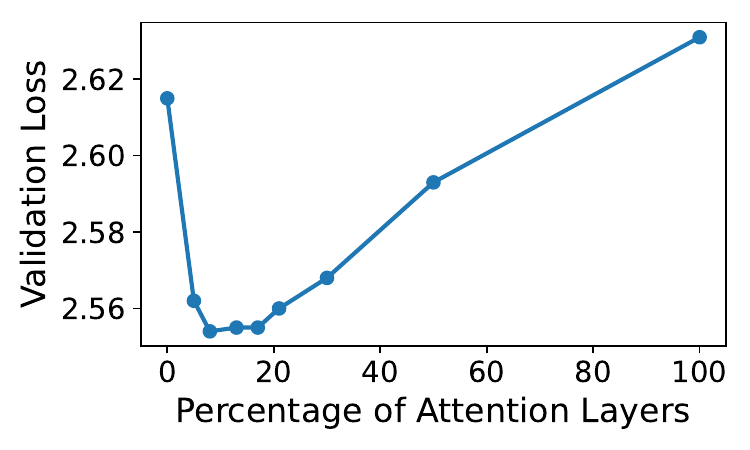}
    \caption{Validation loss versus percentage of attention layers for 130M-parameter hybrid Mamba-Transformer models (24 total layers).}
    \label{fig:att_ratio}
\end{minipage}
\vspace{-0.2in}
\end{wrapfigure}
\newparagraph{Number of Attention and MLP Layers.}
We first study how the number of self-attention and MLP layers in a hybrid model impacts model quality. For these experiments, we train 130M parameter hybrid models with 24 layers, and vary the percentage of the those layers that are attention and that are MLP. As we increase the percentage of these layer types, we evenly distribute them throughout the model, as described in Appendix \ref{appendix:layer-allocation}. We report the validation loss as a function of the attention layer ratio in Figure~\ref{fig:att_ratio}. From these experiments, we discover that validation loss is minimized when roughly 8\% of the layers are self-attention layers. Experiments with 840M parameter models confirm that these findings scale across model sizes. These results are also consistent with those reported by~\cite{dao2024transformers}. After fixing the percentage of attention layers to 8, we vary the percentage of MLP layers between 5 and 50. We conclude that 30\%-50\% of the layers can be MLPs without increasing model loss. In general, we prefer larger MLP layer ratios from an efficiency perspective---with 50\% of the layers set as MLPs, training is 20\% faster than when MLP layers make up only 5\% of the model.

\newparagraph{Position Embeddings.}
We next evaluate whether or not to add Rotary Position Embeddings (RoPE)~\citep{su2024roformer} to every self-attention layer in a hybrid model. For these experiments, we train an 840M-parameter Mamba-Hybrid model on the 1.1T token dataset with and without RoPE, each with an attention layer ratio of 8\%. We use a 4096 sequence length. We then extend these base models to a context length of 16384 through continued pretraining on the longer sequence length for an additional 16B tokens. We experiment with and without adjusting the RoPE base frequency during continued pretraining (continued pretraining with an increase base frequency was introduced by~\citet{xiong2023effective}). Results are reported in Table~\ref{tab:rope_ablation}. The base 840M model trained with RoPE provides a similar accuracy to the model without RoPE, but achieves a lower average accuracy after long context extension (regardless of whether the RoPE base frequency is modified or not). Based on these experiments, as in recent work~\citep{lieber2024jamba}, we opt to ignore RoPE position embeddings for larger-scale hybrid models.

\begin{table}[t]
\scriptsize
\caption{Comparison of 840M parameter hybrid Mamba-Transformer models with and without RoPE position embeddings when using an attention layer ratio of 8\% and training on 1.1T tokens. We train a base model with a sequence length of 4096. We extend this base model to a sequence length of 16384 with continued pretraining for an additional 16B tokens. We find that hybrid models do not need position embeddings, and may perform better without them on long contexts.} 
\label{tab:rope_ablation}
\centering
\begin{tabular}{ccc|ccccccc}
\toprule
Seq. & RoPE & RoPE Freq. Base & WinoGrande & PIQA & HellaSwag & ARC-E & ARC-C & Avg\\
\midrule
\multirow{2}{*}{4K} & no & - & 56.2 & 70.84 & 52.13 & 55.01 & 26.79 & 52.19 \\
  & yes & 10K & 55.8 & 71.71 & 52.3 &	55.43 &	26.71 &	52.39 \\
\midrule
\multirow{3}{*}{16K} & no & - & 56.83 &	72.31 &	53.86 &	56.52 &	27.65 &	\textbf{53.43} \\
& yes & 10K &  54.62 & 71.49 & 53.17 & 56.23 & 27.56 & 52.61 \\
& yes & 500K &  54.7 & 71.38 & 50.42 & 54.29 & 26.79 & 51.52 \\
\bottomrule
\end{tabular}
\end{table}

\newparagraph{Additional Ablation Experiments.}
We also evaluated how the ordering of Mamba/Mamba-2, self-attention, and MLP layers affects the resulting model's natural language abilities (measured with validation loss). When testing, following \citet{park2024can}, we made certain that a Mamba layer appears at the beginning of the architecture---this ensures that the hybrid model can operate without position embeddings, as the first Mamba layer naturally learns to encode the positional information. Our experiments found no significantly better configuration than to evenly distribute self-attention and MLP layers throughout the model, as described in Appendix~\ref{appendix:layer-allocation}. We did not find it necessary to construct hybrid model architectures using a repeated block pattern. We also found that hybrid models can use self-attention layers with Group-Query Attention (GQA)~\citep{ainslie2023gqa} rather than Multi-Head Attention (MHA) with little degradation in model quality (validation perplexity increases $\approx 0.04\%$). Given the decrease in the amount of computation and memory required for inference with GQA compared to MHA, we thus opt to use GQA when training larger-scale hybrid models.

\subsection{Mamba-2-Hybrid 8B}

\newparagraph{Model Architecture and Hyperparameters.}
Based on the study described in Section~\ref{subsec:hybrid_design}, we train an 8B-parameter hybrid SSM-Transformer model with the architecture summarized in Table~\ref{tab:hybrid_arch}. Out of 56 total layers, the hybrid model has 4 (7.1\%) self-attention layers, 24 (42.9\%) Mamba-2 layers, and 28 (50\%) MLP layers. Rather than using a single repeated hybrid block structure, to construct our model we allocate the layers such that 1) a Mamba-2 layer comes first and 2) the attention and MLP layers are evenly distributed throughout the model, as described in Appendix \ref{appendix:layer-allocation}. We use Mamba-2 for the SSM-layers rather than Mamba, as the SSM scan used by Mamba-2 is up to 8$\times$ faster than that of Mamba~\citep{dao2024transformers}. Moreover, our experiments in Section~\ref{subsec:pure_model_eval}, showed that 8B-parameter Mamba-2 models match or exceed 8B-parameter Mamba models on common downstream natural language tasks. For the Mamba-2 layers, we use the same parameters as for our pure Mamba-2 model (Section~\ref{subsec:pure_model_arch}). That is, we use an internal state dimension of 128, eight groups, a head dimension of 64, expansion factor two, and window size of four for convolution. For the attention layers, we use Group Query Attention with eight groups, 32 attention heads, and 128 KV-Channels. For MLP layers, we use a $4\times$ expansion ratio. Throughout the model, we use a hidden dimension of 4096 and GELU activation. We opt to use no explicit position embeddings. For each layer, we include a residual skip connection and RMSNorm before the Mamba-2, self-attention, or MLP block. As for the pure SSM and Transformer models, we do not use Dropout, biases for linear layers, and we use separate parameters for model embeddings and output layer weights (i.e., untied embeddings). We train our Mamba-2-Hybrid 8B on the 1.1T token and 3.5T token datasets using the hyperparameters described in Section~\ref{subsec:pure_model_arch} (i.e., the exact same ones as for the Transformer models and pure SSM models).

\begin{table}[t]
\caption{Summary of the 8B-parameter Mamba-2-Hybrid architecture used throughout the experiments. We design the layer pattern to spread attention and MLP layers evenly throughout the model (Appendix~\ref{appendix:layer-allocation}).}
\label{tab:hybrid_arch}
\centering
\resizebox{\textwidth}{!}{
\begin{tabular}{cccccccccc}
\toprule
Model & Params (B) & \# Layers & Model Dim & Attn. Heads & State Dim. & \# Groups & Pos. Emb. & Seq. Len\\
\midrule
\multirow{4}{*}{Mamba-2-Hybrid} & 8.66 & 56 & 4096 & 32 & 128 & 8 & None & 4096\\
\\
& \multicolumn{8}{c}{Hybrid Layer Pattern (M=Mamba-2, *=Self-Attention, +=MLP)}\\
\cmidrule(lr){2-9}
& \multicolumn{8}{c}{ M+M+M++M+M*+M+M+M+M++M*+M+M+M+M+M*++M+M+M+M+M*+M++M+M+M+}\\
\bottomrule
\end{tabular}
}
\end{table}

\begin{wrapfigure}{R}{0.4\textwidth}
\vspace{-0.25in}
\begin{minipage}{0.4\textwidth}
\centering
    \includegraphics[width=1.0\linewidth]{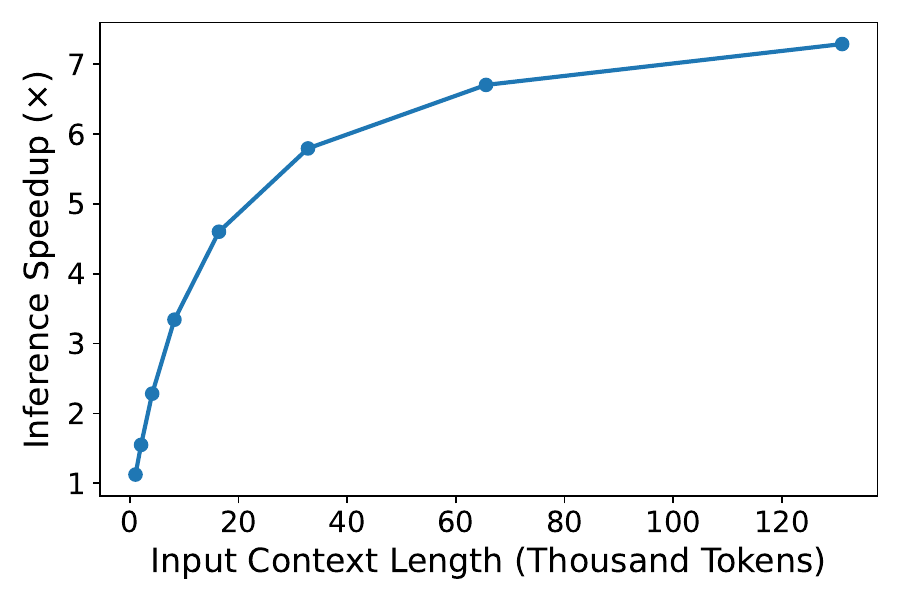}
    \caption{Predicted speedup to generate one token for an 8B-parameter Mamba-2-Hybrid model compared to a Transformer. }
    \label{fig:inference}
\end{minipage}
\vspace{-0.25in}
\end{wrapfigure}
\newparagraph{Training Efficiency.}
We highlight that our Mamba-2-Hybrid model implemented in Megatron-LM can be trained efficiently on thousands of GPUs. To do so, we compare our measured Model Flop Utilization (MFU) with that of Transformers. As in prior work~\citep{korthikanti2205reducing}, we define the MFU as follows: First we define the model FLOPs per second to be the number of FLOPs required to perform a model forward and backward pass divided by the iteration time. We can then define the MFU to be the model FLOPs per second divided by the peak theoretical FLOPs per second of the GPUs used for training. When training on NVIDIA H100 GPUs~\citep{h100}, with a tensor-parallel size of four and data-parallel size of 256 (1024 total GPUs) (micro batch size 4, global batch size 1024), our Mamba-2-Hybrid achieves an MFU of 29.9\%. This can be compared to the 30.7\% MFU of a corresponding 8B parameter Transformer implemented in Megatron-LM and trained with the same parallelization configuration. 

\newparagraph{Inference Speed.}
We also highlight that the hybrid model benefits from the inference-time speedups expected of a pure SSM model compared to a pure Transformer model. In Figure~\ref{fig:inference}, we plot the predicted time to generate one token for the 8B Transformer model over the time for the 8B Mamba-2-Hybrid model using a batch size of 32. For short input context lengths, both models can generate the next token in roughly equivalent time. For long context lengths, however, the hybrid model benefits from its many SSM layers and generates the content nearly 8$\times$ faster than the Transformer. We expect additional inference-time benefits for the hybrid model due to a reduced key-value cache size that should enable Mamba-2-Hybrid to use larger batch sizes than possible with the Transformer model.

\subsection{Empirical Evaluation of Mamba-2-Hybrid}
\subsubsection{Downstream Language Modeling Tasks}
\begin{wrapfigure}{R}{0.4\textwidth}
\vspace{-0.2in}
\begin{minipage}{0.4\textwidth}
\centering
    \includegraphics[width=1.0\linewidth]{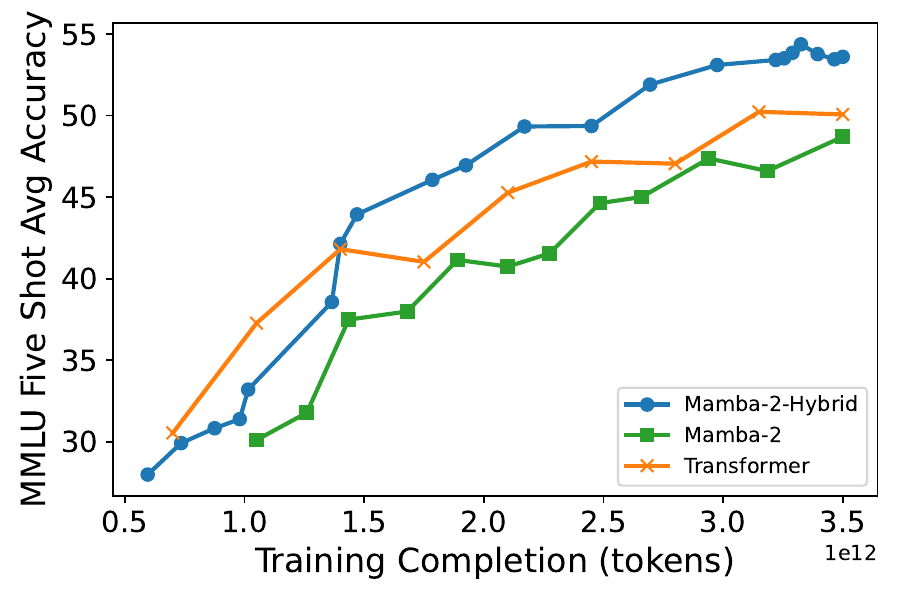}
    \caption{Five-shot MMLU accuracy (standard formulation) for 8B-parameter models trained on 3.5T tokens as a function of training completion.}
    \label{fig:mmlu_vs_training}
\end{minipage}
\vspace{-0.1in}
\end{wrapfigure}
We evaluate the Mamba-2-Hybrid 8B model trained on the 3.5T token dataset using downstream natural language tasks in Table~\ref{tab:three_t_hybrid}. We include comparisons with the pure SSM and Transformer models discussed in Section~\ref{subsec:pure_model_eval}. Remarkably, Mamba-2-Hybrid achieves higher accuracy than the corresponding Transformer on all 12 common natural language tasks we evaluated (Table~\ref{tab:three_t_hybrid}). The average improvement on these tasks compared to the Transformer model is 2.65 points.
Figure~\ref{fig:mmlu_vs_training} shows the behavior of Mamba-2-Hybrid, Mamba-2, and the corresponding Transformer for MMLU accuracy as training progresses. While Mamba-2-Hybrid always leads pure Mamba-2, early in training the hybrid is inferior in this regard to the Transformer. However, once the hybrid matches the Transformer's accuracy at 1.5T tokens it quickly gains and maintains a strong advantage. We believe this motivates further study into the data efficiency and saturation behavior of Mamba-based models.

\begin{table}[t]
\caption{Detailed evaluation results on 12 common natural language tasks comparing an 8B-parameter hybrid model (Mamba-2-Hybrid) with the pure Mamba-2 SSM and Transformer models from Section~\ref{subsec:pure_model_eval} when training for 3.5T tokens. The Mamba-2-Hybrid model achieves the highest overall accuracy and is 2.65 points better than the Transformer on average.} 
\label{tab:three_t_hybrid}
\centering
\resizebox{\textwidth}{!}{
\begin{tabular}{cccccccccccccccc}
\toprule
\multirow{2}{*}{Model} & \multirow{2}{*}{WG} & \multirow{2}{*}{PIQA} & \multirow{2}{*}{HellaSwag} & \multirow{2}{*}{ARC-E} & \multirow{2}{*}{ARC-C} & \multicolumn{2}{c}{MMLU} & \multirow{2}{*}{OpenBook} & \multirow{2}{*}{TruthFul} & \multirow{2}{*}{PubMed} & \multirow{2}{*}{RACE} & \multirow{2}{*}{NQ} & \multirow{2}{*}{SquadV2} & \multirow{2}{*}{Avg}\\
& & & & & & 0-Shot & 5-Shot & & \\
\midrule
Transformer & 69.14 & 78.62	& 75.89 & 73.27 & 43.77 & 45.69 & 50.07 & 42.00 & 35.48 & 69.20 & 39.52 & 15.15 & 53.4 & 53.17\\
Mamba-2 & \textbf{71.59} & \textbf{79.82} & \textbf{77.69} & 75.93 & \textbf{48.12} & 47.25 & 48.7 & \textbf{44.2} & 35.66 & \textbf{75.2} & 37.7 & 17.17 & 51.9 & 54.69\\
\midrule
Mamba-2-Hybrid & 71.27 & 79.65 & \textbf{77.68} & \textbf{77.23} & 47.7 & \textbf{51.46} & \textbf{53.60} & 42.80 & \textbf{38.72} & 69.80 & \textbf{39.71} & \textbf{17.34} & \textbf{58.67} & \textbf{55.82} \\
\bottomrule
\end{tabular}
}
\end{table}

Table~\ref{tab:three_t_hybrid} and Figure~\ref{fig:mmlu_vs_training} show that when training on sufficiently long token horizons (in our case 3.5T tokens), a hybrid model containing Mamba-2, self-attention, and MLP layers can exceed the accuracy of a pure Mamba-2 and a pure Transformer model when averaged over a wide variety of downstream natural language tasks. These results provide exciting evidence for the capability of hybrid models to provide faster LLM inference and greater model quality when compared to Transformers.

\subsubsection{Long-Context Evaluation}
In this section, we evaluate the long-context ability of hybrid SSM-Transformer models by training two Mamba-2-Hybrid 8B extensions---a 16386 (16K) and 32768 (32K) variant---and compare to corresponding extensions of the 8B Transformer. We extend the base models (pretrained using sequence lengths of 4096) to 16K and 32K versions through continued pretraining on the respective larger context lengths. We use full global attention in the four self-attention layers. In this initial study, we use the same underlying data as in our 3.5T dataset. That is, we do not explicitly select a data subset consisting of long documents, but rather use packed sequences to generate 16K and 32K inputs for the model. All long context models are trained for an additional 50B tokens with a learning rate that increases linearly over the first 1.7B tokens and then decays according to cosine annealing thereafter. We use a max learning rate of 3e-5 and minimum learning rate of 3e-6. For the Transformer extensions, we automatically adapt the RoPE base frequency to the longer context lengths using the dynamic NTK scaling described in~\citet{bloc972023ntkrope}.

\begin{table}[t]
\caption{Standard task evaluation results of 16K and 32K sequence length variants of the 8B-parameter Mamba-2-Hybrid and Transformer model trained on 3.5T tokens. These tasks do not require long-context abilities but they show that the accuracy on these tasks does not degrade for the long-context models when compared to the base 4K models.} 
\label{tab:hybrid_long_context1}
\centering
\resizebox{\textwidth}{!}{
\begin{tabular}{cccccccccccccccc}
\toprule
\multirow{2}{*}{Model} & \multirow{2}{*}{WG} & \multirow{2}{*}{PIQA} & \multirow{2}{*}{HellaSwag} & \multirow{2}{*}{ARC-E} & \multirow{2}{*}{ARC-C} & \multicolumn{2}{c}{MMLU} & \multirow{2}{*}{OpenBook} & \multirow{2}{*}{TruthFul} & \multirow{2}{*}{PubMed} & \multirow{2}{*}{RACE} & \multirow{2}{*}{NQ} & \multirow{2}{*}{SquadV2} & \multirow{2}{*}{Avg}\\
& & & & & & 0-Shot & 5-Shot & & \\
\midrule
Transformer-4K & 69.14 & 78.62	& 75.89 & 73.27 & 43.77 & 45.69 & 50.07 & 42.00 & 35.48 & 69.20 & 39.52 & 15.15 & 53.4 & 53.17\\
Mamba-2-Hybrid-4K & \textbf{71.27} & \textbf{79.65} & \textbf{77.68} & \textbf{77.23} & \textbf{47.7} & \textbf{51.46} & \textbf{53.60} & \textbf{42.80} & \textbf{38.72} & \textbf{69.80} & \textbf{39.71} & \textbf{17.34} & \textbf{58.67} & \textbf{55.82} \\
\midrule
Transformer-16K & 70.4 & 78.67 & 76.4 & 74.45 & 44.37 & 47.48 & 51.22 & 42.2 & 36.33 & 69.8 & 38.95 & 14.29 & 55.38 & 53.84\\
Mamba-2-Hybrid-16k & \textbf{71.67} & \textbf{79.92} & \textbf{78.24} & \textbf{77.95} & \textbf{48.12} & \textbf{52.01} & \textbf{54.92} & \textbf{44.60 }& \textbf{37.23} & \textbf{70.00} & \textbf{39.33} & \textbf{18.50} & \textbf{58.99} & \textbf{56.27} \\
\midrule
Transformer-32K & 69.22 & 78.51 & 76.01 & 73.74 & 43.09 & 47.80 & 50.42 & 41.60 & 36.28 & 69.40 & 38.66 & 15.79 & 54.93 & 53.50 \\
Mamba-2-Hybrid-32K & \textbf{71.43} & \textbf{79.54} & \textbf{78.08} & \textbf{78.07} & \textbf{47.70} & \textbf{52.41} & \textbf{55.09} & \textbf{45.40} & \textbf{37.86} & \textbf{71.00} & \textbf{40.10} & \textbf{18.64} & \textbf{57.93} & \textbf{56.4} \\
\bottomrule
\end{tabular}
}
\end{table}

\newparagraph{Results on Standard Short-Context Tasks.}
We first evaluate the 16K and 32K Mamba-2-Hybrid and Transformer models on the 12 standard natural language tasks used above. While these tasks do not require long-context abilities, we aim to check whether model accuracy degrades on common tasks as a result of extending our models to long-context variants. Results are reported in Table~\ref{tab:hybrid_long_context1}. On average, we find no accuracy decrease on these tasks for the long-context variants. In fact, the 16K and 32K models slightly improve compared to the base models which is due to the 16K and 32K models seeing 1.4\% more data. As for the original 4K evaluations, the 16K and 32K Mamba-2-Hybrid is more than 2 points better than the corresponding Transformer models on average.

\newparagraph{Results on Natural Long-Context Tasks.}
We now focus on evaluating the 16K and 32K model extensions on tasks which require natural language reasoning across long contexts. Results when evaluating these models on nine common long-context tasks are shown in Table~\ref{tab:hybrid_long_context2}. In this setting, the base (4K) Mamba-2-Hybrid and Transformer models achieve similar accuracy on most tasks and are more than 6 points better than the pure Mamba-2 model. For both architectures, the 16K and 32K variants improve over the base models by an average of roughly 4 points. This is particularly due to a large accuracy increase on tasks with many long inputs (e.g., NarrativeQA). Comparing the 16K and 32K Mamba-2-Hybrid to the corresponding 16K and 32K Transformer, we observe that the Transformer models separate from the hybrid models on some tasks, particularly Multi-Document Question Answering tasks (e.g., HotpotQA). This leads the 16K and 32K Transformer models to reach approximately one point higher average accuracy than the 16K and 32K Mamba-2-Hybrid models respectively. 

We hypothesize that the hybrid model reaches lower accuracy than the Transformer on these tasks because the SSM layer states are sometimes confused by documents irrelevant to the question (which is unknown until the end of the sequence)---The Muti-Document Question Answering tasks in Table~\ref{tab:hybrid_long_context2} are taken from the LongBench evaluation suite which generates long-context inputs by concatenating the few documents from each task needed to answer the question (e.g., HotpotQA questions contain two paragraphs and then a question which requires knowledge from both) with many random paragraphs sampled from Wikipedia. This confusion could be due to our continued pretraining recipe which simply packs unrelated sequences together to make 16K and 32K inputs, potentially leading the SSM layers to believe separate documents are related when they are not. While this recipe is widely used for Transformers, it may not directly apply to hybrid models. 

Based on our experience evaluating these tasks, we also note that hybrid models may be more sensitive to prompt formatting than Transformer models. As evidence to support this hypothesis, we found that minor prompt modifications could change the results for both models, but more so for the hybrid model. For example, on Musique, prompt modifications led the accuracy for the Mamba-2-Hybrid-4K model to fall in the range [10.63, 16.16]. In contrast, the accuracy for the Transformer was relatively steady, remaining in the range [15.25, 17.68]. We highlight, however, that the prompt format for the majority of the tasks in Table~\ref{tab:hybrid_long_context2} (e.g., the Multi-Document QA tasks, see Section~\ref{sec:preliminaries}) are taken from the LongBench evaluation suite~\citep{bai2023longbench} and have been optimized for Transformer models. As a result of these observations, we believe interesting areas of future work involve further study on the prompt robustness of hybrid models and comparing aligned and instruction-tuned hybrid models to their Transformer counterparts.

\begin{table}[t]
\caption{Natural long-context evaluation results for 8B-parameter base and long-context extensions of Mamba-2-Hybrid and Transformer models trained on 3.5T tokens. While the 4K models are comparable on these tasks, the 16K and 32K Transformer is roughly 1 point better on average when compared to the 16K and 32K Mamba-2-Hybrid. These improvements come largely on Multi-Document Question Answering tasks from LongBench~\citep{bai2023longbench}.} 
\label{tab:hybrid_long_context2}
\centering
\resizebox{\textwidth}{!}{
\begin{tabular}{ccccccccccc}
\toprule
\multirow{3}{*}{Model} & \multicolumn{4}{c}{Single Doc. QA} & \multicolumn{3}{c}{Multi-Doc. QA} & \multicolumn{2}{c}{Few-Shot Learning} & \\
\cmidrule(lr){2-5}
\cmidrule(lr){6-8}
\cmidrule(lr){9-10}
& NarrativeQA & Qasper & MultiFieldQA & QuALITY & HotpotQA & 2WikiMQA & Musique & TREC & TriviaQA & Avg\\
\cmidrule(lr){2-2}
\cmidrule(lr){3-3}
\cmidrule(lr){4-4}
\cmidrule(lr){5-5}
\cmidrule(lr){6-6}
\cmidrule(lr){7-7}
\cmidrule(lr){8-8}
\cmidrule(lr){9-9}
\cmidrule(lr){10-10}
(avg, max) Ctx. Len: & (86K, 506K) & (4.9K, 22K) & (7.2K, 17K) & (6.5K, 9.8K) & (13.3K, 19K) & (7.5K, 17K) & (16.3K, 18K) & (7.1K, 11.9K) & (12.3K, 25K) & \\
\midrule
Mamba-2 & 22.53 & 25.74 & 29.26 & 33.84 & 29.99 & 24.6 & 11.1 & 54 & 77.77 & 34.31 \\
Transformer-4K & 23.44 & 28.51 & 38.39 & \textbf{36.48} & 36.28 & \textbf{33.48} & \textbf{17.68} & 68 & 83.65 & 40.66\\
Mamba-2-Hybrid-4k & \textbf{24.53} & \textbf{28.75} & \textbf{39.01} & 35.6 & 36.24 & 32.91 & 15.24 & \textbf{68.5} & \textbf{86.7} & \textbf{40.83} \\
\midrule
Transformer-16K & 27.51 & 29.71 & \textbf{41.13} & \textbf{39.02} & \textbf{48.61} &\textbf{ 34.87} & 21.42 & \textbf{78.5} & 86.39 & \textbf{45.24} \\
Mamba-2-Hybrid-16k & \textbf{29.76} & \textbf{30.93} & 40.9 & 38.93 & 42.17 & 31.34 & \textbf{23.32} & 74 & \textbf{90.05} & 44.6 \\
\midrule
Transformer-32K & 30.06 & 29.09 & 40.61 & \textbf{39.17} & \textbf{48.2} & \textbf{35.52} & \textbf{24.55} & \textbf{76.5} & 86.43 & \textbf{45.57}\\
Mamba-2-Hybrid-32K & \textbf{31.56} & \textbf{30.55} & \textbf{40.69} & 38.88 & 41.9 & 29.06 & 21.33 & 74.5 & \textbf{89} & 44.16 \\
\bottomrule
\end{tabular}
}
\end{table}

\newparagraph{Results on Synthetic Long-Context Tasks.}
Beyond the natural long-context tasks discussed above, we also evaluate the 16K and 32K hybrid and Transformer extensions on the synthetic tasks in the RULER~\citep{hsieh2024ruler} benchmark suite. These tasks expand upon the basic Needle In A Haystack (NIAH) problem where the model is asked to recall information (the needle) from long inputs of otherwise irrelevant text. RULER also includes tasks which require tracing and aggregating information across the context. For these evaluations, the task context lengths are set to 4K for the base models, 16K for the 16K extensions, and 32K for the 32K models. 

Results on the 13 RULER tasks are shown in Table~\ref{tab:hybrid_long_context_ruler}. Overall, the Mamba-2-Hybrid models show significantly improved NIAH abilities compared to the Transformer models and pure Mamba-2 model. For example, The 16K hybrid model achieves 13 points higher average accuracy on these tasks compared to the 16K Transformer. The long-context Transformer models are particularly challenged by the Variable Tracking (VT) task. This task includes a one shot demonstration of the task in the context and a closer inspection of the model predictions shows that the Transformer tends to directly copy the answer for the in-context example instead of predicting the output of the actual question. This behavior is consistent with prior observations for LWM-7B and Yi-34B models on these tasks~\citep{hsieh2024ruler}. Interestingly, while the hybrid model is generally better on most tasks, the Transformer consistently reaches higher accuracy on Keywords Extraction (KWE). 

We also observe that the hybrid model reaches higher accuracy than the Transformer on the HotpotQA task, which contrasts the behavior in Table~\ref{tab:hybrid_long_context2} when running HotpotQA using the LongBench evaluation suite. As described above, while the latter benchmark constructs long context HotpotQA questions by adding random Wikipedia passages to the relevant information, RULER extends the context length of HotpotQA by adding paragraphs randomly sampled from HotpotQA itself. This slight difference in the distribution used for context creation seems to confuse the hybrid model in one case (i.e., LongBench HotpotQA (Table~\ref{tab:hybrid_long_context2}), but not in the other (i.e., RULER HotpotQA (Table~\ref{tab:hybrid_long_context_ruler}) and provides an interesting area of future study. 

\begin{table}[t]
\caption{Evaluation results of 8B-parameter Mamba-2-Hybrid and Transformer models trained for 3.5T tokens, plus their long-context extensions, on the synthetic RULER long-context benchmark suite~\citep{hsieh2024ruler}. Mamba-2-Hybrid models achieve higher accuracy than Transformers on these tasks, highlighting their improved ability to recall, trace, and aggregate information across long inputs.} 
\label{tab:hybrid_long_context_ruler}
\centering
\resizebox{\textwidth}{!}{
\begin{tabular}{ccccccccccccccc}
\toprule
\multirow{2}{*}{Model} & \multicolumn{5}{c}{Synthetic Tasks} & \multicolumn{8}{c}{Needle-In-A-Haystack Tasks (NIAH)} & \\
\cmidrule(lr){2-6}
\cmidrule(lr){7-14}
& HotpotQA & SquadQA & CWE & VT & KWE & NIAH-1 & NIAH-2 & NIAH-3 & MK-NIAH-1 & MK-NIAH-2 & MK-NIAH-3 & MV-NIAH & MQ-NIAH & Avg\\
\midrule
Mamba-2 & 31.75 & 35.5 & 32.87 & 76.45 & \textbf{76.58} & 100 & 98 & 83 & 40.25 & 13.75 & 5.5 & 35 & 49.12 & 52.14\\
Transformer-4K & 42.25 & 56.5 & \textbf{36.42} & 79.2 & 76.33 & 100 & 100 & 75.25 & \textbf{99.5} & 94 & 58 & 96.56 & 95.06 & 77.62\\
Mamba-2-Hybrid-4k & \textbf{48.75} & 56.5 & 32.2 & \textbf{90.55} & 65.58 & 100 & 100 & \textbf{95.75} & 89.5 & \textbf{95.5} & \textbf{96} & \textbf{97.94} & \textbf{97.62} & \textbf{81.99}\\
\midrule
Transformer-16K & 37.25 & 45 & 3 & 7.25 & \textbf{58.33} & 100 & 99.5 & 75.75 & \textbf{93.75} & 56.5 & 57.5 & 88.5 & 87.94 & 62.33\\
Mamba-2-Hybrid-16k & \textbf{44} & \textbf{48.75} & \textbf{12.88} & \textbf{83.2} & 46.83 & 100 & 100 & \textbf{81.5} & 92 & \textbf{92.25} & \textbf{83} & \textbf{89.81} & \textbf{90.19} & \textbf{74.19}\\
\midrule
Transformer-32K & 35 & 41.5 & 4.42 & 0.35 & \textbf{53.5} & 100 & 99 & 76.5 & 70.75 & 57.75 & 41.25 & 69.69 & \textbf{84.69} & 56.49\\
Mamba-2-Hybrid-32K & \textbf{38.5} & 41.75 & \textbf{8.4} & \textbf{79.9} & 36.5 & 100 & 100 & \textbf{96.75} & \textbf{84} & \textbf{76.5} & \textbf{81.5} & \textbf{84.31} & 80.94 & \textbf{69.93}\\
\bottomrule
\end{tabular}
}
\end{table}

\begin{figure}[t]
  \vspace{0.1in}
  \centering
  \begin{subfigure}[t]{0.32\textwidth}
     \centering
     \includegraphics[width=\textwidth]{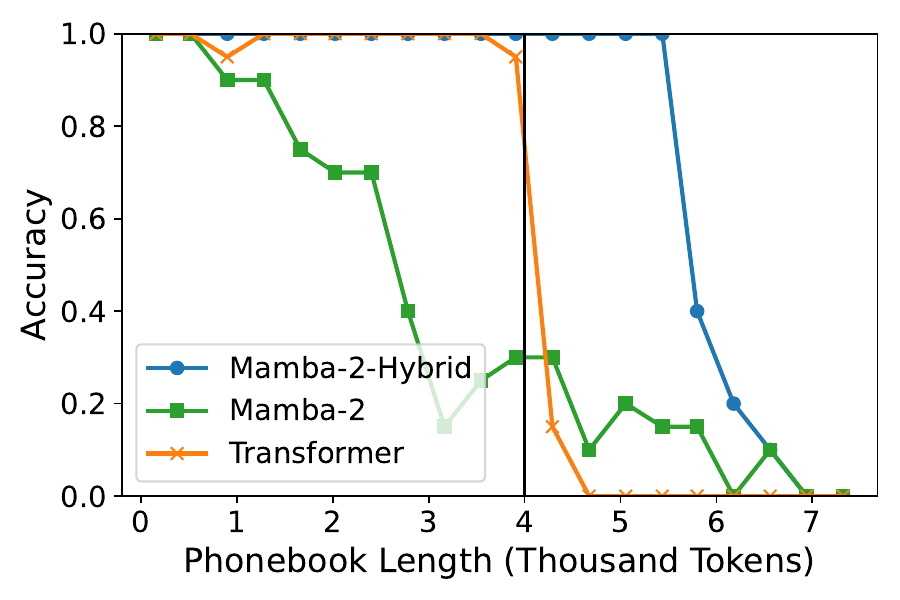}
     \caption{4K base models}
     \label{subfig:phonebook_4k}
  \end{subfigure}
  \hfill
  \begin{subfigure}[t]{0.32\textwidth}
     \centering
     \includegraphics[width=\textwidth]{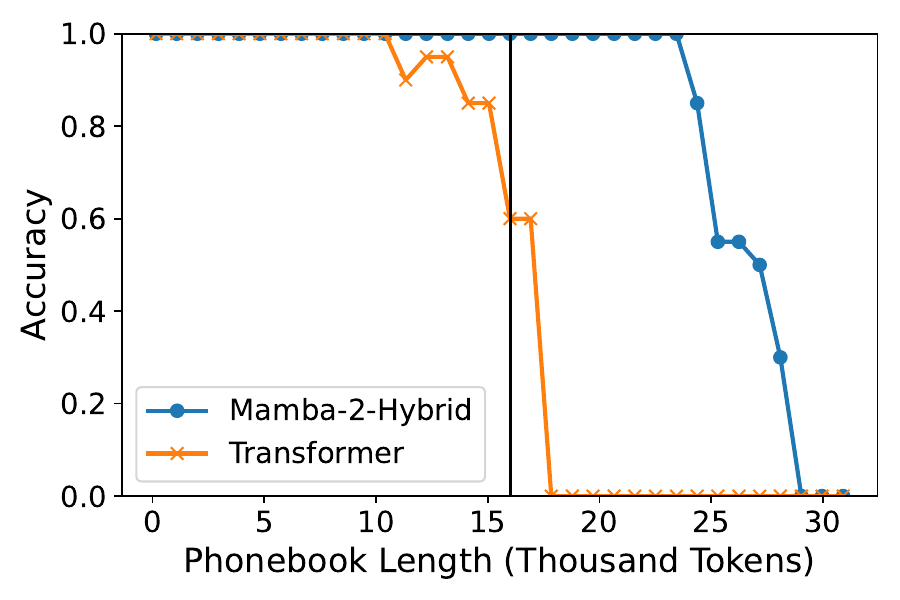}
     \caption{16K models}
     \label{subfig:phonebook_16k}
  \end{subfigure}
  \hfill
  \begin{subfigure}[t]{0.32\textwidth}
     \centering
     \includegraphics[width=\textwidth]{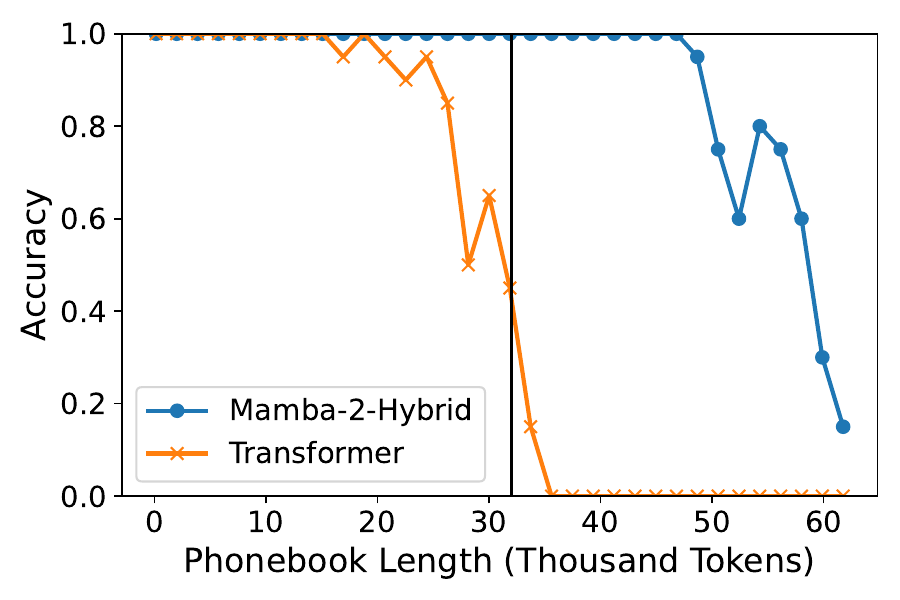}
     \caption{32K models}
     \label{subfig:phonebook_32k}
  \end{subfigure}
  \caption{Phonebook evaluation results for 8B-parameter models trained on 3.5T tokens and their long-context extensions. We use the standard Phonebook formulation (Figure~\ref{fig:phonebook_task}). Mamba-2-Hybrid models can generalize beyond their pretraining sequence length and perform the Phonebook task on contexts longer than pure SSM or Transformer models. \textbf{(a)} Base 4K Mamba-2, Mamba-2-Hybrid, and Transformer model Phonebook evaluations. \textbf{(b)} 16K long-context extensions of the Mamba-2-Hybrid and Transformer model evaluated on Phonebook. \textbf{(c)} Phonebook evaluations for the Mamba-2-Hybrid and Transformer models extended to support 32K sequence lengths.}
  \label{}
\end{figure}

\newparagraph{Results on Copying Tasks: Phonebook.}
Finally, we evaluate the long-context hybrid and Transformer models on the synthetic Phonebook task (Section~\ref{subsec:pure_model_copying}, Figure~\ref{fig:phonebook_task}). We use the standard formulation which tests a model's ability to perform in-context learning and copying from the context. Results for the base 4K models trained on 3.5T tokens are shown in Figure~\ref{subfig:phonebook_4k}. In this Figure, we also include the results for the pure Mamba-2 model trained on 3.5T tokens. As highlighted for pure Mamba models trained on 1.1T tokens (Section~\ref{subsec:pure_model_copying}), the Mamba-2 model is unable to accurately predict the required phone numbers for sequences $>$1000 tokens. In contrast, the Transformer and Mamba-2-Hybrid can do the Phonebook task with near perfect accuracy up to the pretraining context length (4K). In fact, the hybrid model can generalize slightly beyond this sequence length, achieving 100 percent accuracy on Phonebook up to 5.5K tokens. Similar results hold for the long-context models (Figure~\ref{subfig:phonebook_16k} and Figure~\ref{subfig:phonebook_32k}). Both the 16K and 32K Mamba-2-Hybrid extensions can perform the Phonebook task perfectly beyond their trained context length. The long-context Transformer models, however, start to make mistakes as the phone book length approaches their trained context lengths. In Griffin~\citep{de2024griffin}, the authors make similar observations, finding that their Transformer baseline slowly degrades as it approaches the training context length and that hybrid architectures show near-perfect accuracy up to their attention window size. As with the RULER evaluations above, these experiments highlight again the strong ability for hybrid models to perform in-context learning and to retrieve information from a long context.

\begin{wrapfigure}{R}{0.4\textwidth}
\vspace{-0.30in}
\begin{minipage}{0.4\textwidth}
\centering
    \includegraphics[width=1.0\linewidth]{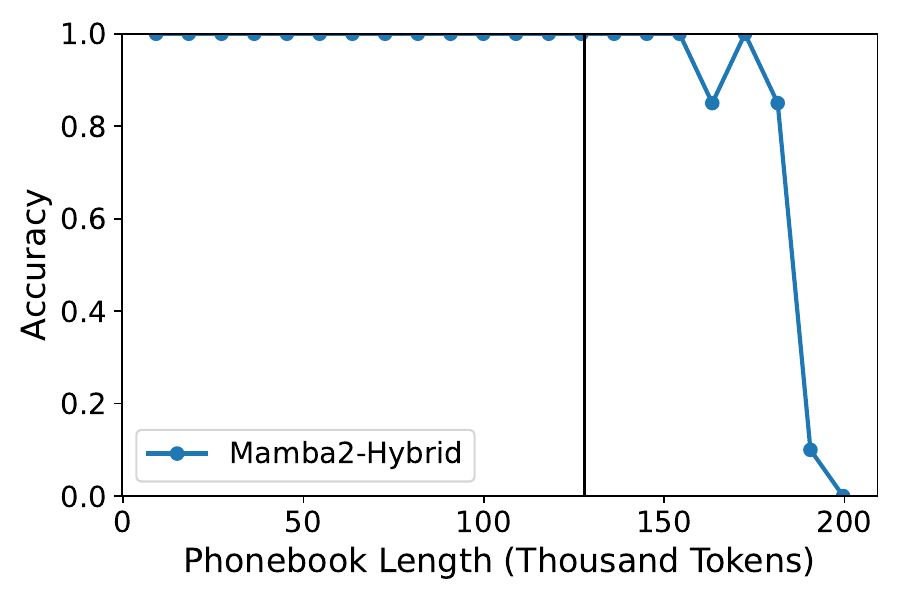}
    \caption{Phonebook accuracy (standard setting) for an 8B Mamba-2-Hybrid trained on 3.5T tokens and extended to 128K sequence length through continued pretraining for 50B tokens.}
    \label{fig:phonebook_128k}
\end{minipage}
\vspace{-0.35in}
\end{wrapfigure}
\newparagraph{A 128K Mamba-2-Hybrid Model.}
While we focused in this section on evaluating 16K and 32K Mamba-2-Hybrid long-context extensions and comparing them to corresponding Transformer models, we now show that the hybrid architecture can extend to context lengths well beyond 32K. We extend the base 4K Mamba-2-Hybrid model to a sequence length of 128K through continued pretraining as described above, using full global attention for the four self-attention layers. This training required only tensor and pipeline parallelism in Megatron-LM to prevent out-of-memory issues. We report results for this model on the Phonebook task in Figure~\ref{fig:phonebook_128k}. As for the 4K, 16K, and 32K Mamba-2-Hybrid models, the 128K model is able to do this task perfectly up to and beyond the sequence length it was trained on. This experiment highlights the promising potential for extending hybrid models to long context lengths.

\newparagraph{Takeaway.}
We have presented a detailed evaluation of long-context 8B-parameter Mamba-2-Hybrid models and compared them with their Transformer counterparts. Overall, the hybrid models match or exceed the long-context capabilities of the Transformers in most tasks. This is particularly true for tasks like Phonebook and the Needle In A Haystack (NIAH) present in the synthetic RULER benchmark. We have identified, however, a few tasks where the hybrid models failed to reach Transformer-level accuracy (e.g., Multi-Document Question Answering in the LongBench evaluation suite). We encourage further research into these settings and into long-context versions of hybrid SSM-Transformer architectures.

\section{Related Work}
Recent work has also introduced Mamba-Attention hybrid models to improve accuracy and efficiency compared to pure Mamba and Transformers. \citet{park2024can} show the limitations of Mamba on in-context learning (ICL) tasks and propose a hybrid model to improve the ICL accuracy. Their experiments, however, are isolated to ICL tasks, and the model size is small (up to 77M parameters). Jamba~\citep{lieber2024jamba} and Zamba~\citep{glorioso2024zamba} train Mamba-Attention hybrid models at 7B scale. 
Both show that their hybrid models significantly improve inference speed and GPU memory compared to other models including Llama~\citep{touvron2023llama} and Mistral-7B~\citep{jiang2023mistral}. Jamba improves the model accuracy and efficiency by adding Mixture-of-Experts (MoE), which increases the total model capacity (52B total parameters) but not its active parameters. They compare their hybrid architecture with pure Mamba and Transformer models on four standard and three long-context tasks, but only using 1.3B parameter models trained for 250B tokens, or 7B parameter models trained for 50B tokens. Zamba introduces a shared attention module and uses an annealing phase during training with high-quality datasets, which boosts the quality of their hybrid model. We focus on combining Mamba, attention, and MLP layers into hybrid models for direct comparison with Transformer baselines at larger scales ($>$7B parameters and $>$1T tokens). 

Other recent work introduces hybrid models that mix either linear RNNs or convolutions with attention. 
\citet{de2024griffin} introduce a hybrid model that blends gated linear recurrences with local (sliding window) attention and show that the hybrid model can improve next token prediction latency with increasing context length. They train 1B parameter models on 8K sequence lengths for long-context modeling. \citet{arora2023zoology} report that a simple convolution-attention hybrid model outperforms pure attention in multi-query associative recall problems while reducing total FLOPs. 
Several additional works add SSM layers to the Transformer architecture to increase accuracy: \citet{saon2023diagonal} use SSM layers together with Transformers to improve speech recognition quality. \citet{pilault2024block} combine an SSM and a block-wise Transformer at every layer. They show improved perplexity and generalization capabilities for longer sequences (up to 65K). Their model is scaled up to 1.3B parameters.
We note that all the hybrid models mentioned above have manually designed architectures and place an MLP layer (if they use MLP) after each attention layer, similar to Transformers. Our study, however, finds that a specific hybrid architecture design or pattern is not required. Instead, the relative proportions of each type of hybrid component appears to be the key factor that determines the quality of the model.

\section{Conclusion}
To address the question of whether SSM models can match the accuracy of Transformers at larger training budgets, in this report we presented a direct experimental comparison between 8B-parameter Mamba, Mamba-2, Mamba-2-Hybrid, and Transformer models trained on up to 3.5T tokens. Our experiments showed that pure SSM models match or exceed the capabilities of their Transformer counterparts on most downstream tasks but are challenged by tasks that require context-based information retrieval (e.g., copying) and in-context learning. We also showed that hybrid SSM-Transformer models (Mamba-2-Hybrid) reach higher accuracy than Transformers on all common benchmarks we evaluated. Further, these hybrid models continue to show strong capabilities compared to Transformers when extended to 16K and 32K contexts. Based on these results, we are encouraged by the potential for SSM-based models to deliver inference-time speedups without accuracy degradation compared to Transformer models.
We look forward to future work focusing on how hybrid models can make use of the large ecosystem of frameworks, methods, and libraries currently tailored to the large-scale training and inference of Transformers.

\printbibliography

@inproceedings{gu2021efficiently,
  title={{Efficiently Modeling Long Sequences with Structured State Spaces}},
  author={Gu, Albert and Goel, Karan and Re, Christopher},
  booktitle={International Conference on Learning Representations},
  year={2021}
}

@article{gu2023mamba,
  title={{Mamba: Linear-time Sequence Modeling with Selective State Spaces}},
  author={Gu, Albert and Dao, Tri},
  journal={arXiv preprint arXiv:2312.00752},
  year={2023}
}

@inproceedings{dao2024transformers,
  title={{Transformers are {SSM}s: Generalized Models and Efficient Algorithms Through Structured State Space Duality}},
  author={Dao, Tri and Gu, Albert},
  booktitle={International Conference on Machine Learning (ICML)},
  year={2024}
}

@inproceedings{hendrycks2020measuring,
  title={{Measuring Massive Multitask Language Understanding}},
  author={Hendrycks, Dan and Burns, Collin and Basart, Steven and Zou, Andy and Mazeika, Mantas and Song, Dawn and Steinhardt, Jacob},
  booktitle={International Conference on Learning Representations},
  year={2020}
}

@article{park2024can,
  title={{Can Mamba Learn How to Learn? A Comparative Study on In-Context Learning Tasks}},
  author={Park, Jongho and Park, Jaeseung and Xiong, Zheyang and Lee, Nayoung and Cho, Jaewoong and Oymak, Samet and Lee, Kangwook and Papailiopoulos, Dimitris},
  journal={arXiv preprint arXiv:2402.04248},
  year={2024}
}

@article{xiong2023effective,
  title={{Effective Long-context Scaling of Foundation Models}},
  author={Xiong, Wenhan and Liu, Jingyu and Molybog, Igor and Zhang, Hejia and Bhargava, Prajjwal and Hou, Rui and Martin, Louis and Rungta, Rashi and Sankararaman, Karthik Abinav and Oguz, Barlas and others},
  journal={arXiv preprint arXiv:2309.16039},
  year={2023}
}

@article{su2024roformer,
  title={{Roformer: Enhanced Transformer with Rotary Position Embedding}},
  author={Su, Jianlin and Ahmed, Murtadha and Lu, Yu and Pan, Shengfeng and Bo, Wen and Liu, Yunfeng},
  journal={Neurocomputing},
  volume={568},
  pages={127063},
  year={2024},
  publisher={Elsevier}
}

@article{bloc972023ntkrope,
  title={{NTK-aware Scaled RoPE allows LLaMA models to have Extended (8k+) Context Size Without any Fine-tuning and Minimal Perplexity Degradation}},
  author={bloc97},
  year={2023},
  url={https://www.reddit.com/r/LocalLLaMA/comments/14lz7j5/ntkaware_
scaled_rope_allows_llama_models_to_have}
}

@article{lieber2024jamba,
  title={{Jamba: A Hybrid Transformer-mamba Language Model}},
  author={Lieber, Opher and Lenz, Barak and Bata, Hofit and Cohen, Gal and Osin, Jhonathan and Dalmedigos, Itay and Safahi, Erez and Meirom, Shaked and Belinkov, Yonatan and Shalev-Shwartz, Shai and others},
  journal={arXiv preprint arXiv:2403.19887},
  year={2024}
}

@article{glorioso2024zamba,
  title={{Zamba: A Compact 7B SSM Hybrid Model}},
  author={Glorioso, Paolo and Anthony, Quentin and Tokpanov, Yury and Whittington, James and Pilault, Jonathan and Ibrahim, Adam and Millidge, Beren},
  journal={arXiv preprint arXiv:2405.16712},
  year={2024}
}

@article{de2024griffin,
  title={{Griffin: Mixing Gated Linear Recurrences with Local Attention for Efficient Language Models}},
  author={De, Soham and Smith, Samuel L and Fernando, Anushan and Botev, Aleksandar and Cristian-Muraru, George and Gu, Albert and Haroun, Ruba and Berrada, Leonard and Chen, Yutian and Srinivasan, Srivatsan and others},
  journal={arXiv preprint arXiv:2402.19427},
  year={2024}
}

@article{arora2023zoology,
  title={{Zoology: Measuring and Improving Recall in Efficient Language Models}},
  author={Arora, Simran and Eyuboglu, Sabri and Timalsina, Aman and Johnson, Isys and Poli, Michael and Zou, James and Rudra, Atri and R{\'e}, Christopher},
  journal={arXiv preprint arXiv:2312.04927},
  year={2023}
}

@article{pilault2024block,
  title={{Block-state Transformers}},
  author={Pilault, Jonathan and Fathi, Mahan and Firat, Orhan and Pal, Chris and Bacon, Pierre-Luc and Goroshin, Ross},
  journal={Advances in Neural Information Processing Systems},
  volume={36},
  year={2024}
}

@article{touvron2023llama,
  title={{Llama 2: Open Foundation and Fine-tuned Chat Models}},
  author={Touvron, Hugo and Martin, Louis and Stone, Kevin and Albert, Peter and Almahairi, Amjad and Babaei, Yasmine and Bashlykov, Nikolay and Batra, Soumya and Bhargava, Prajjwal and Bhosale, Shruti and others},
  journal={arXiv preprint arXiv:2307.09288},
  year={2023}
}

@article{jiang2023mistral,
  title={{Mistral 7B}},
  author={Jiang, Albert Q and Sablayrolles, Alexandre and Mensch, Arthur and Bamford, Chris and Chaplot, Devendra Singh and Casas, Diego de las and Bressand, Florian and Lengyel, Gianna and Lample, Guillaume and Saulnier, Lucile and others},
  journal={arXiv preprint arXiv:2310.06825},
  year={2023}
}

@inproceedings{saon2023diagonal,
  title={{Diagonal State Space Augmented Transformers for Speech Recognition}},
  author={Saon, George and Gupta, Ankit and Cui, Xiaodong},
  booktitle={ICASSP 2023-2023 IEEE International Conference on Acoustics, Speech and Signal Processing (ICASSP)},
  pages={1--5},
  year={2023},
  organization={IEEE}
}

@article{vaswani2017attention,
  title={{Attention is All You Need}},
  author={Vaswani, Ashish and Shazeer, Noam and Parmar, Niki and Uszkoreit, Jakob and Jones, Llion and Gomez, Aidan N and Kaiser, {\L}ukasz and Polosukhin, Illia},
  journal={Advances in Neural Information Processing Systems},
  volume={30},
  year={2017}
}

@article{achiam2023gpt,
  title={{GPT-4 Technical Report}},
  author={Achiam, Josh and Adler, Steven and Agarwal, Sandhini and Ahmad, Lama and Akkaya, Ilge and Aleman, Florencia Leoni and Almeida, Diogo and Altenschmidt, Janko and Altman, Sam and Anadkat, Shyamal and others},
  journal={arXiv preprint arXiv:2303.08774},
  year={2023}
}

@article{bahdanau2014neural,
  title={{Neural Machine Translation by Jointly Learning to Align and Translate}},
  author={Bahdanau, Dzmitry and Cho, Kyunghyun and Bengio, Yoshua},
  journal={arXiv preprint arXiv:1409.0473},
  year={2014}
}

@article{tay2022efficient,
  title={{Efficient Transformers: A Survey}},
  author={Tay, Yi and Dehghani, Mostafa and Bahri, Dara and Metzler, Donald},
  journal={ACM Computing Surveys},
  volume={55},
  number={6},
  pages={1--28},
  year={2022},
  publisher={ACM New York, NY}
}

@article{shoeybi2019megatron,
  title={{Megatron-LM: Training Multi-billion Parameter Language Models using Model Parallelism}},
  author={Shoeybi, Mohammad and Patwary, Mostofa and Puri, Raul and LeGresley, Patrick and Casper, Jared and Catanzaro, Bryan},
  journal={arXiv preprint arXiv:1909.08053},
  year={2019}
}

@inproceedings{narayanan2021efficient,
  title={{Efficient Large-scale Language Model Training on GPU Clusters using Megatron-LM}},
  author={Narayanan, Deepak and Shoeybi, Mohammad and Casper, Jared and LeGresley, Patrick and Patwary, Mostofa and Korthikanti, Vijay and Vainbrand, Dmitri and Kashinkunti, Prethvi and Bernauer, Julie and Catanzaro, Bryan and others},
  booktitle={Proceedings of the International Conference for High Performance Computing, Networking, Storage and Analysis},
  year={2021}
}

@article{korthikanti2205reducing,
  title={{Reducing Activation Recomputation in Large Transformer Models}},
  author={Korthikanti, Vijay and Casper, Jared and Lym, Sangkug and McAfee, Lawrence and Andersch, Michael and Shoeybi, Mohammad and Catanzaro, Bryan},
  journal={arXiv preprint arXiv:2205.05198},
  year={2022}
}

@article{parmar2024nemotron,
  title={{Nemotron-4 15B Technical Report}},
  author={Parmar, Jupinder and Prabhumoye, Shrimai and Jennings, Joseph and Patwary, Mostofa and Subramanian, Sandeep and Su, Dan and Zhu, Chen and Narayanan, Deepak and Jhunjhunwala, Aastha and Dattagupta, Ayush and others},
  journal={arXiv preprint arXiv:2402.16819},
  year={2024}
}

@article{kudo2018sentencepiece,
  title={{Sentencepiece: A Simple and Language Independent Subword Tokenizer and Detokenizer for Neural Text Processing}},
  author={Kudo, Taku and Richardson, John},
  journal={arXiv preprint arXiv:1808.06226},
  year={2018}
}

@article{hsieh2024ruler,
  title={{RULER: What's the Real Context Size of Your Long-Context Language Models?}},
  author={Hsieh, Cheng-Ping and Sun, Simeng and Kriman, Samuel and Acharya, Shantanu and Rekesh, Dima and Jia, Fei and Ginsburg, Boris},
  journal={arXiv preprint arXiv:2404.06654},
  year={2024}
}

@inproceedings{yang2018hotpotqa,
  title={{HotpotQA: A Dataset for Diverse, Explainable Multi-hop Question Answering}},
  author={Yang, Zhilin and Qi, Peng and Zhang, Saizheng and Bengio, Yoshua and Cohen, William and Salakhutdinov, Ruslan and Manning, Christopher D},
  booktitle={Proceedings of the 2018 Conference on Empirical Methods in Natural Language Processing},
  pages={2369--2380},
  year={2018}
}

@inproceedings{ho2020constructing,
  title={{Constructing A Multi-hop QA Dataset for Comprehensive Evaluation of Reasoning Steps}},
  author={Ho, Xanh and Nguyen, Anh-Khoa Duong and Sugawara, Saku and Aizawa, Akiko},
  booktitle={Proceedings of the 28th International Conference on Computational Linguistics},
  pages={6609--6625},
  year={2020}
}

@article{trivedi2022musique,
  title={{MuSiQue: Multihop Questions via Single-hop Question Composition}},
  author={Trivedi, Harsh and Balasubramanian, Niranjan and Khot, Tushar and Sabharwal, Ashish},
  journal={Transactions of the Association for Computational Linguistics},
  volume={10},
  pages={539--554},
  year={2022},
  publisher={MIT Press One Broadway, 12th Floor, Cambridge, Massachusetts 02142, USA~…}
}

@article{kovcisky2018narrativeqa,
  title={{The NarrativeQA Reading Comprehension Challenge}},
  author={Ko{\v{c}}isk{\`y}, Tom{\'a}{\v{s}} and Schwarz, Jonathan and Blunsom, Phil and Dyer, Chris and Hermann, Karl Moritz and Melis, G{\'a}bor and Grefenstette, Edward},
  journal={Transactions of the Association for Computational Linguistics},
  volume={6},
  pages={317--328},
  year={2018},
  publisher={MIT Press One Rogers Street, Cambridge, MA 02142-1209, USA journals-info~…}
}

@inproceedings{dasigi2021dataset,
  title={{A Dataset of Information-Seeking Questions and Answers Anchored in Research Papers}},
  author={Dasigi, Pradeep and Lo, Kyle and Beltagy, Iz and Cohan, Arman and Smith, Noah A and Gardner, Matt},
  booktitle={Proceedings of the 2021 Conference of the North American Chapter of the Association for Computational Linguistics: Human Language Technologies},
  pages={4599--4610},
  year={2021}
}

@inproceedings{joshi2017triviaqa,
  title={{TriviaQA: A Large Scale Distantly Supervised Challenge Dataset for Reading Comprehension}},
  author={Joshi, Mandar and Choi, Eunsol and Weld, Daniel S and Zettlemoyer, Luke},
  booktitle={Proceedings of the 55th Annual Meeting of the Association for Computational Linguistics (Volume 1: Long Papers)},
  pages={1601--1611},
  year={2017}
}

@inproceedings{li2002learning,
  title={{Learning Question Classifiers}},
  author={Li, Xin and Roth, Dan},
  booktitle={COLING 2002: The 19th International Conference on Computational Linguistics},
  year={2002}
}

@article{bai2023longbench,
  title={{LongBench: A Bilingual, Multitask Benchmark for Long Context Understanding}},
  author={Bai, Yushi and Lv, Xin and Zhang, Jiajie and Lyu, Hongchang and Tang, Jiankai and Huang, Zhidian and Du, Zhengxiao and Liu, Xiao and Zeng, Aohan and Hou, Lei and Dong, Yuxiao and Tang, Jie and Li, Juanzi},
  journal={arXiv preprint arXiv:2308.14508},
  year={2023}
}

@article{shaham2022scrolls,
  title={{Scrolls: Standardized Comparison over Long Language Sequences}},
  author={Shaham, Uri and Segal, Elad and Ivgi, Maor and Efrat, Avia and Yoran, Ori and Haviv, Adi and Gupta, Ankit and Xiong, Wenhan and Geva, Mor and Berant, Jonathan and others},
  journal={arXiv preprint arXiv:2201.03533},
  year={2022}
}

@article{clark2018think,
  title={{Think you have Solved Question Answering? Try ARC, the AI2 Reasoning Challenge}},
  author={Clark, Peter and Cowhey, Isaac and Etzioni, Oren and Khot, Tushar and Sabharwal, Ashish and Schoenick, Carissa and Tafjord, Oyvind},
  journal={arXiv preprint arXiv:1803.05457},
  year={2018}
}

@article{zellers2019hellaswag,
  title={{HellaSwag: Can a Machine Really Finish your Sentence?}},
  author={Zellers, Rowan and Holtzman, Ari and Bisk, Yonatan and Farhadi, Ali and Choi, Yejin},
  journal={arXiv preprint arXiv:1905.07830},
  year={2019}
}

@article{lee2019latent,
  title={{Latent Retrieval for Weakly Supervised Open Domain Question Answering}},
  author={Lee, Kenton and Chang, Ming-Wei and Toutanova, Kristina},
  journal={arXiv preprint arXiv:1906.00300},
  year={2019}
}

@article{mihaylov2018can,
  title={{Can a Suit of Armor Conduct Electricity? A New Dataset for Open Book Question Answering}},
  author={Mihaylov, Todor and Clark, Peter and Khot, Tushar and Sabharwal, Ashish},
  journal={arXiv preprint arXiv:1809.02789},
  year={2018}
}

@article{jin2019pubmedqa,
  title={{PubMedQA: A Dataset for Biomedical Research Question Answering}},
  author={Jin, Qiao and Dhingra, Bhuwan and Liu, Zhengping and Cohen, William W and Lu, Xinghua},
  journal={arXiv preprint arXiv:1909.06146},
  year={2019}
}

@article{lai2017race,
  title={{RACE: Large-scale ReAding Comprehension Dataset From Examinations}},
  author={Lai, Guokun and Xie, Qizhe and Liu, Hanxiao and Yang, Yiming and Hovy, Eduard},
  journal={arXiv preprint arXiv:1704.04683},
  year={2017}
}

@article{rajpurkar2018know,
  title={{Know what you don't Know: Unanswerable Questions for SQuAD}},
  author={Rajpurkar, Pranav and Jia, Robin and Liang, Percy},
  journal={arXiv preprint arXiv:1806.03822},
  year={2018}
}

@article{lin2021truthfulqa,
  title={{TruthfulQA: Measuring How Models Mimic Human Falsehoods}},
  author={Lin, Stephanie and Hilton, Jacob and Evans, Owain},
  journal={arXiv preprint arXiv:2109.07958},
  year={2021}
}

@article{sakaguchi2021winogrande,
  title={{WinoGrande: An Adversarial Winograd Schema Challenge at Scale}},
  author={Sakaguchi, Keisuke and Bras, Ronan Le and Bhagavatula, Chandra and Choi, Yejin},
  journal={Communications of the ACM},
  volume={64},
  number={9},
  pages={99--106},
  year={2021},
  publisher={ACM New York, NY, USA}
}

@inproceedings{bisk2020piqa,
  title={{PIQA: Reasoning about Physical Commonsense in Natural Language}},
  author={Bisk, Yonatan and Zellers, Rowan and Gao, Jianfeng and Choi, Yejin and others},
  booktitle={Proceedings of the AAAI Conference on Artificial Intelligence},
  volume={34},
  number={05},
  pages={7432--7439},
  year={2020}
}

@misc{eval-harness,
  author       = {Gao, Leo and Tow, Jonathan and Abbasi, Baber and Biderman, Stella and Black, Sid and DiPofi, Anthony and Foster, Charles and Golding, Laurence and Hsu, Jeffrey and Le Noac'h, Alain and Li, Haonan and McDonell, Kyle and Muennighoff, Niklas and Ociepa, Chris and Phang, Jason and Reynolds, Laria and Schoelkopf, Hailey and Skowron, Aviya and Sutawika, Lintang and Tang, Eric and Thite, Anish and Wang, Ben and Wang, Kevin and Zou, Andy},
  title        = {{A Framework for Few-shot Language Model Evaluation}},
  month        = 12,
  year         = 2023,
  publisher    = {Zenodo},
  version      = {v0.4.0},
  url          = {https://zenodo.org/records/10256836}
}

@article{brown2020language,
  title={{Language Models are Few-shot Learners}},
  author={Brown, Tom and Mann, Benjamin and Ryder, Nick and Subbiah, Melanie and Kaplan, Jared D and Dhariwal, Prafulla and Neelakantan, Arvind and Shyam, Pranav and Sastry, Girish and Askell, Amanda and others},
  journal={Advances in Neural Information Processing Systems},
  volume={33},
  pages={1877--1901},
  year={2020}
}

@article{shazeer2020glu,
  title={{GLU Variants Improve Transformer}},
  author={Shazeer, Noam},
  journal={arXiv preprint arXiv:2002.05202},
  year={2020}
}

@article{ba2016layer,
  title={{Layer Normalization}},
  author={Ba, Jimmy Lei and Kiros, Jamie Ryan and Hinton, Geoffrey E},
  journal={arXiv preprint arXiv:1607.06450},
  year={2016}
}

@article{zhang2019root,
  title={{Root Mean Square Layer Normalization}},
  author={Zhang, Biao and Sennrich, Rico},
  journal={Advances in Neural Information Processing Systems},
  volume={32},
  year={2019}
}

@article{hendrycks2016gaussian,
  title={{Gaussian Error Linear Units (GELUs)}},
  author={Hendrycks, Dan and Gimpel, Kevin},
  journal={arXiv preprint arXiv:1606.08415},
  year={2016}
}

@article{ainslie2023gqa,
  title={{GQA: Training Generalized Multi-Query Transformer Models from Multi-head Checkpoints}},
  author={Ainslie, Joshua and Lee-Thorp, James and de Jong, Michiel and Zemlyanskiy, Yury and Lebr{\'o}n, Federico and Sanghai, Sumit},
  journal={arXiv preprint arXiv:2305.13245},
  year={2023}
}

@misc{h100,
  title = {{NVIDIA H100 Tensor Core GPU}},
  howpublished = {\url{https://www.nvidia.com/en-us/data-center/h100/}},
  author={NVIDIA},
  year={2023}
}

@article{jelassi2024repeatafterme,
  title={{Repeat After Me: Transformers are Better than State Space Models at Copying}},
  author={Jelassi, Samy and Brandfonbrener, David and Kakade, Sham M and Malach, Eran},
  journal={arXiv preprint arXiv:2402.01032},
  year={2024}
}

\appendix
\section{Hybrid Layer Allocation Algorithm}
\label{appendix:layer-allocation}

Although we are able to specify, and experiment with, an arbitrary sequence of Mamba, self-attention, and MLP layers in our hybrid models, by default we use the allocation algorithm described in Algorithm~\ref{alg:allocate_layers}. This algorithm first attempts to place any self-attention layers such that the intervening runs of contiguous Mamba layers are as equal in length as possible, while also beginning and ending the layer sequence with a run of Mamba layers. Then, any MLP layers are evenly distributed throughout the sequence while not replacing any self-attention layers. The MLP layers are biased away from the start of the sequence so that the layer sequence begins with a Mamba layer (if there are any Mamba layers) and ends with an MLP layer (if there are any MLP layers).

\begin{algorithm}
\caption{Hybrid Layer Allocation}
\label{alg:allocate_layers}
\begin{algorithmic}[1]
    \State \textbf{Input:} total\_layers\_count, target\_attention\_ratio, target\_mlp\_ratio
    \State \textbf{Output:} layer\_type\_list
    \State attention\_layers\_count $\gets$ \text{round}(total\_layers\_count * target\_attention\_ratio)
    \State mamba\_layers\_count $\gets$ total\_layers\_count - attention\_layers\_count
    \State mamba\_sections\_count $\gets$ attention\_layers\_count + 1
    \State mamba\_section\_length $\gets$ mamba\_layers\_count / mamba\_sections\_count

    \State layer\_type\_list $\gets$ \text{array of Symbols.MAMBA of size total\_layers\_count}
    \State x $\gets$ mamba\_section\_length
    \For{l in 0 to total\_layers\_count - 1}
        \If{x < 0.5}
            \State layer\_type\_list[l] $\gets$ \text{Symbols.ATTENTION}
            \State x $\gets$ x + mamba\_section\_length
        \Else
            \State x $\gets$ x - 1
        \EndIf
    \EndFor

    \State mlp\_layers\_count $\gets$ \text{round}(total\_layers\_count * target\_mlp\_ratio)
    \If{mlp\_layers\_count > 0}
        \State mamba\_layers\_count $\gets$ mamba\_layers\_count - mlp\_layers\_count
        \State mamba\_to\_mlp\_ratio $\gets$ mamba\_layers\_count / mlp\_layers\_count

        \State x $\gets$ mamba\_to\_mlp\_ratio
        \For{l in 0 to total\_layers\_count - 1}
            \If{layer\_type\_list[l] == Symbols.MAMBA}
                \If{x < 0.5}
                    \State layer\_type\_list[l] $\gets$ \text{Symbols.MLP}
                    \State x $\gets$ x + mamba\_to\_mlp\_ratio
                \Else
                    \State x $\gets$ x - 1
                \EndIf
            \EndIf
        \EndFor
    \EndIf
\end{algorithmic}
\end{algorithm}

Table \ref{tab:layer-patterns} provides examples of some layer patterns generated by Algorithm \ref{alg:allocate_layers}.

\begin{table}[H]
\centering
\begin{tabularx}{\textwidth}{|c|c|c|X|}
\hline
TLC & ATT & MLP & Pattern \\
\hline
24  & 0.00 & 0.00 & \texttt{MMMMMMMMMMMMMMMMMMMMMMMM} \\
24  & 0.08 & 0.00 & \texttt{MMMMMMM*MMMMMMMM*MMMMMMM} \\
24  & 0.17 & 0.00 & \texttt{MMMM*MMMM*MMMM*MMMM*MMMM} \\
24  & 0.08 & 0.30 & \texttt{MM+MM+M*M+MMM+MM*+MM+MM+} \\
24  & 0.08 & 0.50 & \texttt{M+M+M++*M+M+M+M+*M++M+M+} \\
24  & 0.50 & 0.50 & \texttt{+*+*+*+*+*+**+*+*+*+*+*+} \\
48  & 0.08 & 0.50 & \texttt{M+M+M++M+*M+M+M+M++*M+M+M+M+*M++M+M+M+*M+M++M+M+} \\
56  & 0.08 & 0.50 & \texttt{M+M+M++M+M*+M+M+M+M++M*+M+M+M+M+M*++M+M+M+M+M*+M++M+M+M+} \\
\hline
\end{tabularx}
\caption{Some examples of hybrid layer patterns generated by Algorithm \ref{alg:allocate_layers}. TLC=total\_layer\_count, ATT=target\_attention\_ratio, MLP=target\_mlp\_ratio. In the pattern, M=Mamba, *=Self-Attention, and +=MLP.}
\label{tab:layer-patterns}
\end{table}

\null
\vfill

\end{document}